\documentclass[runningheads]{llncs}

\usepackage{eccv}

\usepackage{eccvabbrv}
\usepackage{algorithm}
\usepackage{algorithmic}
\usepackage{makecell}
\usepackage{mwe}
\usepackage{wrapfig}
\usepackage{longtable}
\usepackage{graphicx}
\usepackage{booktabs}
\usepackage{bbm}
\usepackage[dvipsnames]{xcolor}
\usepackage{color}
\usepackage{booktabs} 
\usepackage{color, colortbl}

\usepackage[accsupp]{axessibility}  

\usepackage[pagebackref,breaklinks,colorlinks,citecolor=eccvblue]{hyperref}

\usepackage{orcidlink}

\begin{document}

\title{Evaluating Text-to-Image Generative Models: An Empirical Study on Human Image Synthesis} 

\titlerunning{Evaluating Text-to-Image Generative Models}

\author{Muxi Chen$^*$\inst{1} \and
Yi Liu$^*$\inst{1} \and
Jian Yi\inst{1} \and
Changran Xu\inst{1} \and
Qiuxia	Lai\inst{2} \and
HongliangWang\inst{2} \and
Tsung-Yi Ho\inst{1}\and
Qiang Xu\inst{1}}

\authorrunning{Chen et al.}

\institute{The Chinese University of Hong Kong \\ \email{mxchen21,yliu22,jyi1,crxu1,tyho,qxu@cse.cuhk.edu.hk} \and
Communication University of China \\ \email{qxlai,hl\_w@cuc.edu.cn} }

\maketitle
\begin{abstract}
In this paper, we present an empirical study introducing a nuanced evaluation framework for text-to-image (T2I) generative models, applied to human image synthesis. Our framework categorizes evaluations into two distinct groups: first, focusing on image qualities such as aesthetics and realism, and second, examining text conditions through concept coverage and fairness. We introduce an innovative aesthetic score prediction model that assesses the visual appeal of generated images and unveils the first dataset marked with low-quality regions in generated human images to facilitate automatic defect detection. Our exploration into concept coverage probes the model's effectiveness in interpreting and rendering text-based concepts accurately, while our analysis of fairness reveals biases in model outputs, with an emphasis on gender, race, and age. While our study is grounded in human imagery, this dual-faceted approach is designed with the flexibility to be applicable to other forms of image generation, enhancing our understanding of generative models and paving the way to the next generation of more sophisticated, contextually aware, and ethically attuned generative models. Code and data, including the dataset annotated with defective areas, are available at \href{https://github.com/cure-lab/EvaluateAIGC}{https://github.com/cure-lab/EvaluateAIGC}.

\end{abstract}    
\section{Introduction}
\label{sec:intro}


\begin{figure*}[t]
  \centering
   \includegraphics[width=0.94\linewidth]{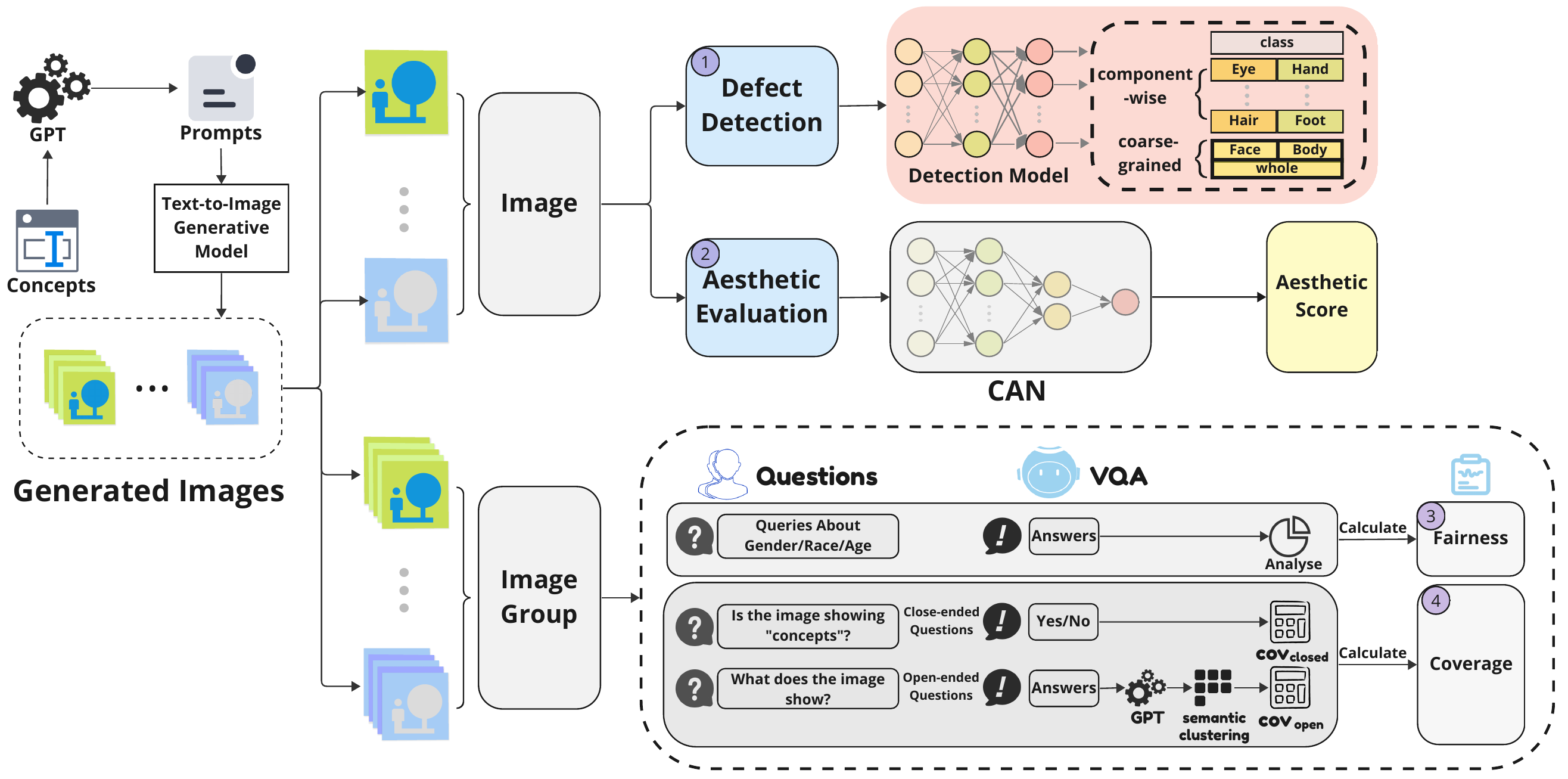}
   \caption{An overview of our framework. Given a T2I generative model for evaluation, we first generate prompts from diverse human-related concepts. Then, we use the T2I generative model to generate images. Each generated image undergoes a dual evaluation process: 1) its aesthetic appeal is assessed using our novel aesthetic score prediction model, CAN, and 2) its realism is evaluated by models trained on our annotated defect dataset. For all images associated with a specific concept, 3) our VQA-based methodologies are employed to discern the absence of biases, and 4) our proposed concept coverage metrics are utilized to assess the fidelity of concept depiction.}
   \label{fig:overview}
\end{figure*}

Recently, the field of text-to-image (T2I) generation has witnessed a paradigm shift, primarily driven by the rapid advancements in diffusion models~\cite{ho2020denoising,song2020denoising,ramesh2022hierarchical,rombach2022high}. Models such as SDXL~\cite{podell2023sdxl} and Midjourney\footnote{\url{https://www.midjourney.com} \\$^*$ Equal Contribution} have demonstrated unparalleled capabilities in creating detailed and diverse images, marking a new era in digital creativity and visual expression. Despite the advancements, these models still cannot guarantee to consistently produce high-quality images that align accurately with intended text prompts.

This inconsistency primarily arises from the gap between the models' training goals, which focus on reconstruction, and their practical applications with prompt-based generation, leading to two significant challenges. Firstly, there is often a lack of realism, \textit{e.g.}, in capturing intricate details such as hand gestures in human image synthesis. Secondly, there is an inadequacy in accurately interpreting and visualizing the text prompts, essential for generating images that are both visually appealing and contextually relevant. These challenges necessitate a thorough understanding and precise evaluation of T2I generative models to fully grasp their capabilities and identify areas for enhancement. However, existing evaluation metrics, \textit{e.g.}, the Inception Score (IS)~\cite{salimans2016improved} and Fréchet Inception Distance (FID)~\cite{heusel2017gans}, only provide a limited view of generative model performance.

To address these limitations, this paper introduces a comprehensive evaluation framework that segments evaluations of T2I generative models into two primary dimensions: \textbf{image quality assessment}, emphasizing aesthetics and realism; and \textbf{text condition evaluation}, focusing on concept coverage and fairness. In the realm of image quality, we have developed an innovative aesthetic score prediction model that quantitatively assesses the aesthetic quality of generated human images. Additionally, we have created the first dataset annotated with low-quality regions in generated human images, facilitating the training of models for automatic defect detection. For text condition evaluation, we assess concept coverage to gauge T2I generative models' adherence to specific prompts. We also explore potential biases in model outputs, particularly focusing on gender, race, and age, to evaluate fairness in the generated images. An overview of our framework is shown in Figure \ref{fig:overview}.

We deploy our framework to evaluate a diverse range of recent T2I generative models, including SD1.5, SD2.1~\cite{rombach2022high}, SDXL, and Midjourney. The results from our evaluations align closely with human assessments, highlighting the effectiveness and robustness of our proposed framework. While our study primarily concentrates on human image synthesis, the adaptable nature of our framework shows great promise for a broader spectrum of applications across various image generation domains, laying the groundwork for the future development of sophisticated, contextually aware, and ethically aligned generative models. 

\section{Related works}
\label{sec:related}

Recent advancements in T2I generative models, \textit{e.g.}, Stable Diffusion~\cite{rombach2022high}, GLIDE \cite{nichol2021glide}, Imagen~\cite{saharia2022photorealistic}, DALL·E~\cite{ramesh2022hierarchical}, and Midjourney, have significantly enhanced their performance and applicability across various fields. The assessment of T2I generative models typically focuses on two crucial aspects: image quality and image-text alignment.

\textbf{Image quality: }When evaluating image quality for T2I generative models, commonly used  metrics are IS and FID. IS assesses the quality and diversity of generated images using a pre-trained Inception network, while FID measures the similarity between the distributions of real and generated images in feature space. Both metrics offer valuable insights, yet they primarily assess image quality at the distribution level rather than individual images. 

Regarding individual image assessment~\cite{zhang2021comprehensive}, image aesthetic assessment is an active field. Recent research increasingly emphasizes that the style information (\textit{e.g.}, the distinctions between portraits and group photos) is a key factor in aesthetic scoring~\cite{kong2016photo, he2022rethinking, yi2023towards}, and has adopted style-aware approaches to evaluate images more effectively. For instance, TANet~\cite{he2022rethinking} derives rules for predicting aesthetics based on an understanding of image styles, demonstrating notable success on benchmark datasets. Similarly, SAAN~\cite{yi2023towards} employs a combination of style-specific and generic aesthetic features to assess the artistic quality of images. While effective to some extent, the generalization capability of these methods is limited as they rely on predefined style annotations in the given dataset. 

\textbf{Image-text alignment:} The most famous metric for image-text alignment is CLIPScore~\cite{hessel2021clipscore, radford2021learning}, initially designed for image captioning. However, CLIPScore can yield inflated results for models optimized in the CLIP space~\cite{nichol2021glide}. Moreover, based on our experiments, we find that CLIPScore does not accurately reflect image-text alignment, often exhibiting misalignment with human observations. Some studies use human assessments to evaluate the semantic coherence between generated images and input prompts~\cite{otani2023toward}, but relying solely on human assessments can be time-consuming, expensive, and impractical for conducting extensive statistical analyses. 
{
Recently, TIFA~\cite{hu2023tifa} employs visual question answering (VQA) to measure the single-image alignment of a generated image to its text input by looking into aspects such as object and counting. In contrast, our work delves into statistically evaluating text conditions through the lens of concept coverage and fairness, which is applied to a collection of generated images featuring the desired concept.

\textbf{Combination:} 
ImageReward\cite{xu2024imagereward} introduces a unified evaluation framework for T2I generative models that considers various facets such as image-text alignment, fidelity, and harmlessness. Nevertheless, it falls short of meeting individualized evaluation requirements across different dimensions and lacks interpretability for each aspect. Our work complements ImageReward by providing a disentangled evaluation process that allows for a detailed assessment of various aspects, namely, aesthetic attributes, the authenticity of human body components, concept coverage, and fairness considerations. }

\section{Image quality assessment}
\label{sec:method1}
Assessing the quality of generated images provides valuable insights into a model's generative capability. In this section, we present our innovative methods for evaluating the aesthetic appeal and realism of the generated images.

\subsection{Design of aesthetic score prediction model}

\begin{figure}[]
  \centering
   \includegraphics[width=0.8\linewidth]{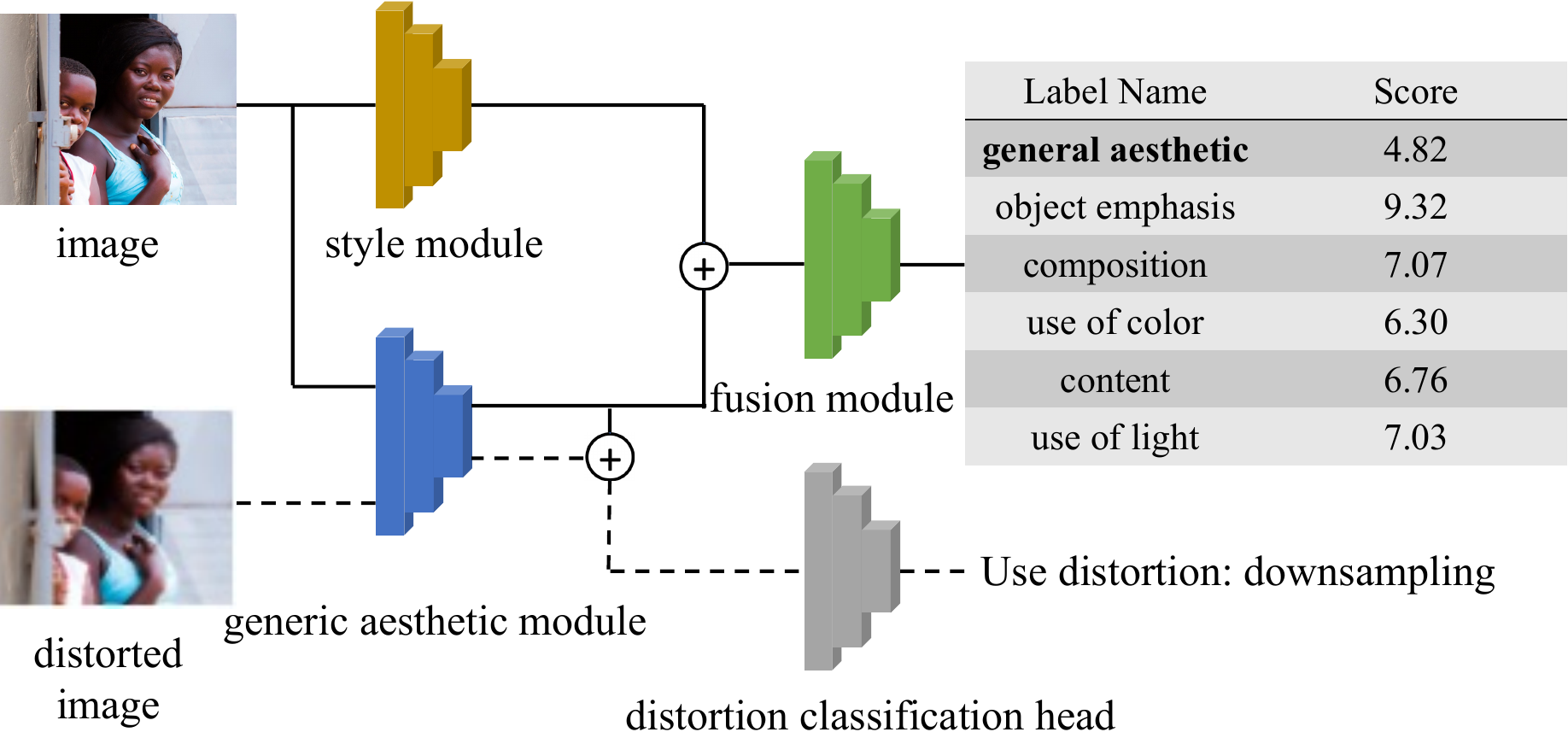}
   \caption{An overview of CAN. CAN predicts scores for the general aesthetic feeling and specific aesthetic attributes of an image. In the training process, the overall model is trained to predict aesthetic scores. Additionally, the generic aesthetic module is trained to predict the distortion applied to an image. $\oplus$ denotes concatenation.}
   \label{fig:aes_model}
\end{figure}

Aesthetics measures the generative model's capacity to generate visually appealing images, implying its understanding of aesthetic factors such as object emphasis, composition, and use of color and light. It is important to note that in our case of human image synthesis, aesthetic assessment should not be confused with measuring the subjective beauty level of an individual~\cite{diamant2019beholder}.

In this paper, we propose a CLIP-based Aesthetic score prediction Network (CAN) for predicting aesthetic scores. Unlike existing methods that rely on pre-defined styles in aesthetic datasets for learning style extraction, CAN leverages the image understanding capability of CLIP~\cite{radford2021learning}, which enables better generalization than existing techniques (\textit{e.g.}, TANet and SAAN) and makes it well-suited for evaluating diverse generated images.

CAN comprises three modules: 1) a style module that uses CLIP to extract the style of an image; 2) a generic aesthetic module that extracts aesthetic features such as color, light, and textures; and 3) a fusion module that fuses the features from the style and generic aesthetic modules to predict aesthetic scores. The architecture of CAN is designed to deliberately separate the processes of describing and assessing aesthetic attributes. Specifically, the generic aesthetic module provides a detailed description of the aesthetic attributes present in an image. In parallel, the style module formulates assessment rules that contribute to the overall evaluation of the image's aesthetic qualities. The fusion module synthesizes these two distinct perspectives. The outputs of CAN are scores for an image's general aesthetic feeling and several aesthetic attributes, including object emphasis, composition, use of color, content, and use of light. Figure~\ref{fig:aes_model} provides an overview of the model architecture.

\textbf{Style module.} While recent aesthetic prediction networks commonly incorporate style extraction, their reliance on predefined style annotations in current aesthetic datasets poses limitations when evaluating the diverse range of styles in generated images. To address this issue, we propose a more general approach to style identification by leveraging the comprehensive knowledge encoded in the pre-trained CLIP model. Trained on a vast dataset of image-text pairs, CLIP excels in image captioning tasks, making it well-suited for extracting image styles.

\textbf{Generic aesthetic module.} The generic aesthetic module aims to extract various generic aesthetic attributes from the image, such as light, color, and composition. To help the model learn to extract these attributes, we design a distortion prediction task~\cite{yi2023towards}, the specifics of which are elaborated in the training scheme. In our experiments, we use ResNet-34~\cite{he2016deep} as the backbone.

\textbf{Fusion module.} We use concatenation and self-attention layers to fuse features from the style and generic aesthetic modules. Subsequently, a series of fully connected layers are employed to predict scores for both the general aesthetic feeling and specific aesthetic attributes. 

\textbf{Training scheme.} We train CAN with two objectives. Firstly, the overall model is trained on the AVA dataset~\cite{murray2012ava} using regression loss to predict aesthetic scores. Simultaneously, we train the generic aesthetic module on distortion prediction tasks. Specifically, the generic aesthetic module takes two images sequentially: the original image from the training set and a distorted version of the same image. The features obtained from the generic aesthetic module for both images are concatenated and passed through a multi-layer perceptron (MLP) to classify the distortion (only one) applied to the image. This objective ensures that the features of the generic aesthetic module contain information such as light, color, and composition of the image.
The final loss $L$ is the sum of the regression loss $L_{reg}$ and the distortion classification loss $L_{cls}$:
\begin{equation}
\centering
\begin{aligned}
    &L_\text{reg} = (F_\text{CAN}(x) - s)^2,\\
    &L_\text{cls} = -\log(P_{d}(F_{cls}(F_\text{gen}(x) \oplus F_\text{gen}(x^{d})))), \\
    &L = L_\text{reg} + L_\text{cls}.
\end{aligned}
\end{equation}
Here, $x$ and $x^{d}$ are the input image and the distorted image, respectively, $F_{CAN}$ denotes the whole model, $F_{gen}$ is the generic aesthetic module, $F_{cls}$ is the distortion classification head, and $s$ is the corresponding aesthetic score. We use $P_d$ to represent the predicted probability of the distortion label $d$.

\subsection{Evaluation of the proposed aesthetic model}

%

\begin{table}
  \centering
  \caption{Comparison of our proposed CAN and TANet in aesthetic score prediction. Both models are trained on the AVA dataset. 
  }
  \begin{tabular}{c|c|ccc}
    \toprule
    Method & Metric & AVA & PARA & TAD66K \\ \midrule
        
    TANet & SRCC & \textbf{0.755} & 0.721 & 0.640 \\
        & Rank Acc. &  \textbf{0.780}  & 0.614  &  0.420     \\ \hline
    CAN & SRCC & 0.754 & \textbf{0.751} & \textbf{0.643}\\
        & Rank Acc. & \textbf{0.780} & \textbf{0.681}   & \textbf{0.431}  \\
    
    \bottomrule
  \end{tabular}
  \label{tab:aesthetic_perf}
  \vspace{-4mm}
\end{table}

\textbf{Experiment setup.} To assess the prediction performance of the generic aesthetic feeling by our CAN model, we train CAN using the training set of the AVA dataset. And we use Spearman’s rank correlation coefficient (SRCC)~\cite{he2022rethinking} and pair-wise rank accuracy as the evaluation metrics. SRCC measures the strength and direction of the monotonic relationship between the predicted aesthetic scores and the ground truth scores, while pair-wise rank accuracy represents the percentage of correctly ranked pairs by the model within the dataset. 

\begin{figure*}[]
  \centering

   \includegraphics[width=1\linewidth]{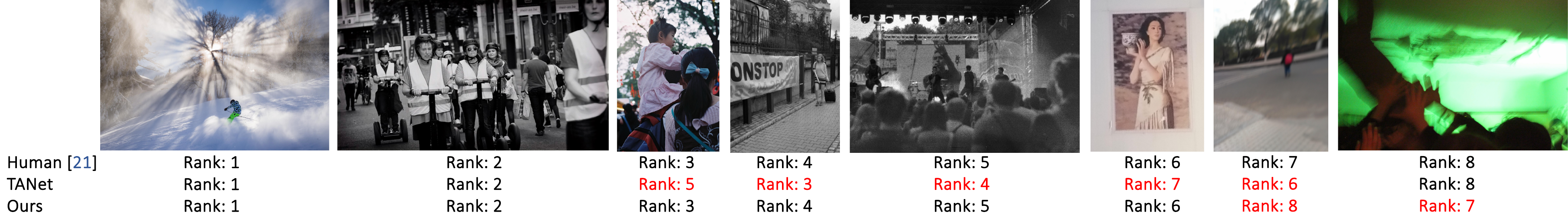}
   \caption{
   The rankings of eight human images by CAN and TANet compared with the ground truth. Red color indicates incorrect rankings.}
   \label{fig:aes_rank}
\end{figure*}

\textbf{Results.} Table~\ref{tab:aesthetic_perf} presents a performance comparison between CAN and the current state-of-the-art (SOTA) model of the AVA dataset, TANet~\cite{he2022rethinking}. Our findings reveal that CAN achieves comparable performance to TANet on the AVA test set. However, when applied to other datasets like TAD66K~\cite{he2022rethinking} and PARA~\cite{yang2022personalized}, CAN demonstrates superior performance in zero-shot prediction. This can be attributed to the enhanced generalization ability of CAN, which leverages the broader knowledge encoded in CLIP. The superior generalization ability of CAN makes it well-suited for evaluating diverse generated images. We provide a visualization of the ranking of human images in the PARA dataset in Figure~\ref{fig:aes_rank}. {In Appendix 1.4, we include an ablation study that examines the effectiveness of the distortion prediction task as well as each module of CAN. }

\subsection{A human dataset annotated with defective areas}

Recent generative models still make mistakes when generating human images, such as distorted faces and hands, which significantly impact the realism of the generated images. The defect rate of generated images serves as a crucial metric for evaluating the ability of generative models to generate realistic images. 

In this paper, we tackle the challenge of identifying defects in generated human images by formulating it as a classification task. Our objective is to train a model that can accurately distinguish between well-painted human images and those with defects. Due to the lack of existing studies in this area, we take the initiative to construct the first dataset that includes annotations for defective human components within generated images.

\textbf{Data collection.} First, we design a set of diverse human-related prompts, encompassing various characteristics such as age, height, weight, gender, race, \textit{etc}. For a complete list of these prompts, please refer to Appendix 2.1. Next, we use SDXL, an advanced T2I generative model that is publicly available, as the foundation model for data generation. We collect 10,000 generated samples. To facilitate model training, we also collect and annotate real human images from three popular datasets: CLIP images from AISegment\footnote{\url{https://www.aisegment.cn}}, CelebAMask-HQ~\cite{CelebAMask-HQ}, and DeepFashion~\cite{liuLQWTcvpr16DeepFashion}. In total, the dataset contains 79,908 images.

\begin{figure}[htbp]
    \centering
    \begin{minipage}[t]{.5\textwidth}
        \centering
        \vspace{0pt}
        \includegraphics[width=.9\textwidth]{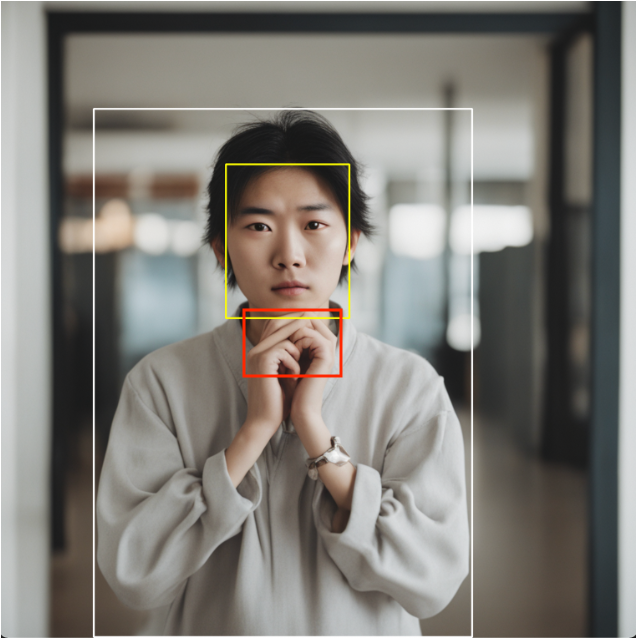}
    \end{minipage}%
    \begin{minipage}[t]{.5\textwidth}
        \centering
        \vspace{0pt}
        \resizebox{0.76\textwidth}{!}{
            \begin{tabular}{lccc}
    \toprule 
    \small\textbf{Class Name} & \small\textbf{Good} & \small\textbf{Bad} & \small\textbf{Invisible} \\ 
    \midrule 
    \small Eye & \small\checkmark &  &  \\ 
    \small Nose & \small\checkmark & & \\ 
    \small Mouth & \small\checkmark &  &  \\ 
    \small Hair & \small\checkmark &  &  \\ 
    \small Cheek & \small\checkmark & & \\ 
    \midrule
    \small Face & \small\checkmark & & \\
    \midrule
    \small Hand &  & \small\checkmark & \\ 
    \small Arm & \small\checkmark & & \\ 
    \small Foot & & & \small\checkmark \\ 
    \small Leg &  & & \small\checkmark \\ 
    \small Trunk & \small\checkmark & & \\ 
    \midrule 
    \small Body & & \small\checkmark & \\ 
    \midrule 
    \midrule 
    \small Whole & & \small\checkmark & \\ 
    \bottomrule 
\end{tabular}


        }
    \end{minipage}
    \caption{Bounding boxes and label assignments for a generated human image. The face box is highlighted in yellow, while the human body box is depicted in white. The table shows annotations of both component-wise and coarse-grained labels. The red bounding box denotes areas with defects.}
    \label{fig:ann_example}
    \vspace{-10pt}
\end{figure}

\textbf{Annotations.} 
We hire a professional team for the annotation of generated samples. The annotations contain two parts. Firstly, the bounding boxes for the face and human body are annotated in each image. Secondly, we assign labels from the set $\{good, bad, invisible\}$ to ten distinct human body components: eye, nose, mouth, hair, cheek, hand, arm, foot, leg, and trunk. The label $good$ signifies a realistic appearance, while $bad$ indicates a defect or anomaly, such as an unusual number of fingers for the hand component.
Apart from labels for individual components, we also merge labels for face components, main body components (without face components), and all components into coarse-grained labels \textit{face}, \textit{body} and \textit{whole}, respectively. When merging these labels, we consider the coarse-grained label as \textit{bad} if any of the labels to be merged is \textit{bad}. Conversely, we consider the coarse-grained label as \textit{invisible} if all the labels to be merged are \textit{invisible}. For the remaining cases, we assign \textit{good} to the coarse-grained label. An example of annotations is presented in Figure~\ref{fig:ann_example}. {Due to space limitations, the annotation details of real images and the label distribution of the dataset are provided in Appendix 2.2 \& 2.3.}

\subsection{Exploration in automatic defect identification}

{\textbf{Can existing metrics or models identify defects?} In Appendix 2.4 \& 2.5, we study the effectiveness of established metrics, such as IS and FID, alongside human detection and pose estimation models in defect identification. Our investigation reveals that IS and FID are inadequate in identifying defects in generated images, as they primarily analyze image distributions and may overlook the defects. Attempts to utilize confidence scores from models like RTMDet-l~\cite{lyu2022rtmdet} and RTMPose-l~\cite{jiang2023rtmpose} for defect identification encounter challenges, including unclear thresholds and sensitivity to variations in component size.}

\textbf{Model Training.} The aforementioned observations underscore the imperative need to train models specifically designed for defect identification. To this end, we train a ViT-based classification model~\cite{he2022masked}, which has dominated many classification tasks in recent years.

In this paper, we explore two tasks based on our dataset. Firstly, we train models for identifying defects in individual components, such as classifying hands as \textit{good}, \textit{invisible}, or \textit{bad}. These models are referred to as component-wise evaluation models. Secondly, we utilize the coarse-grained label \textit{face} to train a model for distinguishing between well-painted and poorly drawn faces. This model is denoted as the face evaluation model.

Testing on generated images, our results reveal an average accuracy of 88\% for the component-wise evaluation models and 86\% for the face evaluation model. While acknowledging that these models' performance falls short of perfection, our exploration demonstrates the feasibility of this task, thus marking it as a potential avenue for future research. {We present more details of the experiment, an evaluation of the models' generalization ability and an in-depth failure analysis in Appendix 2.6-8.}
\section{Evaluating text conditions}
\label{sec:method2}
This section delves into the examination of text conditions, specifically focusing on concept coverage and fairness within the context of human image synthesis. Concept coverage assesses the adherence of generated images to the given human-related prompts. Fairness evaluation focuses on dimensions including gender, race, and age, aiming to uncover any potential biases in the models.

\subsection{Concept coverage}
Concept coverage involves evaluating a model's ability to understand and faithfully represent a given concept in the generated images. 
CLIPScore~\cite{hessel2021clipscore, radford2021learning}, a metric widely employed to measure image-text similarity using CLIP embeddings, serves as a commonly used method to assess the alignment between the provided prompts and the generated images. However, the threshold for determining alignment or misalignment can vary across different prompts. Thus, it cannot provide definitive assessments, thereby restricting its direct applicability to concept coverage.

In contrast, VQA, which also relies on the embeddings of large vision-language models, offers a more definitive approach by providing conclusive answers. Based on VQA, we propose two novel metrics for assessing concept coverage: \textit{\textbf{closed-ended coverage}} ($\text{cov}_{\text{closed}}$) and \textit{\textbf{open-ended coverage}} ($\text{cov}_{\text{open}}$). The calculations for $\text{cov}_{\text{closed}}$ and $\text{cov}_{\text{open}}$ share a similar procedure but differ in the type of questions posed to the VQA model. In this section, we focus on human-related concepts, especially the actions and interactions of humans with their environments and other objects. However, the proposed metrics can be readily adapted for various concepts in other domains.

\textbf{Metrics.} Given a concept $\mathbf{c}$, we first generate a bunch of images $\mathcal{X}=\{\mathbf{x}_{1},\ldots,\mathbf{x}_{n}\}$ using prompts that features the concept. To obtain $\text{cov}_{\text{closed}}$, we incorporate the intended concept $\mathbf{c}$ into the input questions for VQA, generating closed-ended questions like 
"Is the person $\mathbf{c}$ in the image?" or "Is the picture depicting that a human is $\mathbf{c}$?". By questioning the VQA model on all generated images, $\text{cov}_{\text{closed}}$ can be calculated as:
\begin{equation}
    \text{cov}_{\text{closed}} = \frac{\sum_{i=1}^{n} \mathbbm{1}(ans(\mathbf{x}_{i}) == \text{"yes"})}{|\mathcal{X}|}
\end{equation}
where $|\mathcal{X}|$ denotes the total number of images generated, $ans(\mathbf{x}_{i})$ denotes the output answer (either "yes" or "no" in this scenario) for $\mathbf{x}_{i}$, and $\mathbbm{1}(\cdot)$ is an indicator function which returns 1 if $ans(\mathbf{x}_{i}) == \text{"yes"}$ and 0 otherwise.

To obtain $\text{cov}_{\text{open}}$, we use open-ended questions like "What action is the person performing in the image?" or "Can you describe the activity being carried out by the person?". For each generated image $\mathbf{x}_{i}$, we query the model $T$ times and obtain a corresponding answer set $\mathbf{a}_{i}=\{a_{i1},\ldots, a_{iT}\}$. Drawing inspiration from uncertainty estimation techniques in the language domain~\cite{lin2023generating,kuhn2023semantic}, we perform semantic clustering on $\mathbf{a}_{i}$, which group answers with the same semantic meaning. Then, we use the answer in the largest cluster as the final answer for $\mathbf{x}_{i}$. We also compute the semantic entropy of the answers using the cluster distribution. Intuitively, when $\mathbf{x}_{i}$ presents a concept clearly, the answers within $\mathbf{a}_{i}$ are expected to exhibit a high degree of semantic coherence, resulting in a low semantic entropy. The procedure for conducting semantic clustering and calculating semantic entropy is summarized in Algorithm \ref{Algo:Sementic-Clustering}. We use ChatGPT~\cite{ouyang2022training} to determine the semantic equivalence of two answers. For more details, please refer to Appendix 3.2.

With the semantic entropy ($\text{se}_{i}$) and final answer ($\text{ans}_{i}$) for each $\mathbf{x}_{i}$, $\text{cov}_{\text{open}}$ is calculated as:
\begin{equation}
    \text{cov}_{\text{open}} = \frac{\sum_{i=1}^{n} \mathbbm{1}(\text{se}_{i} \leq \delta \; \textit{and} \; \text{sem\_eq} (\text{ans}_{i}, \mathbf{c}))}{|\mathcal{X}|}
\end{equation}
where $\delta$ is a predefined threshold and $\text{sem\_eq}(\cdot)$ denotes the $\text{semantic\_equivalent}(\cdot)$ in Algorithm \ref{Algo:Sementic-Clustering}. Notably, $\delta$ does not vary across prompts and can be adjusted based on the user's preference: a lower $\delta$ implies a more stringent definition of alignment between the prompt and the generated image. We set $\delta\!=\!0.8$.

\begin{algorithm}[htbp]
    \caption{Semantic Clustering and Semantic Entropy}
    \label{Algo:Sementic-Clustering}
    \textbf{Input:} an answer set $\mathbf{a}_{i} = \{a_{i1}, \ldots, a_{iT}\}$ \\
    \textbf{Output:} \textit{semantic\_set\_ids} that assigns each $a_{ij}$ a semantic cluster id; \textit{semantic\_set\_counts} that records the number of elements for each cluster id; \textit{semantic\_entropy} that evaluates the uncertainty of $\mathbf{x}_{i}$; \textit{final\_answer} that stores the voted answer.
    \begin{algorithmic}[1]
    \STATE $\text{\textit{semantic\_set\_ids}} = \{\}$
    \FOR{$a_{ij} \in \mathbf{a}_{i}$}
        \STATE $\text{\textit{semantic\_set\_ids}}[a_{ij}] = j$
    \ENDFOR
    \FOR{$m$ in $\text{range}(T)$}
        \FOR{$n$ in $\text{range}(m+1, T)$}
            \IF{semantic\_equivalent($a_{im}$, $a_{in}$) is True}
                \STATE $\text{\textit{semantic\_set\_ids}}[a_{in}] = \text{\textit{semantic\_set\_ids}}[a_{im}]$
            \ENDIF
        \ENDFOR
    \ENDFOR
    \STATE $\text{\textit{semantic\_set\_counts}} = \text{Counter}(\text{\textit{semantic\_set\_ids}})$
    \STATE Calculate \textit{semantic\_entropy} based on cluster probabilities obtained from \textit{semantic\_set\_counts}
    \STATE Return \textit{final\_answer} belonging to the largest cluster
    \end{algorithmic}
\end{algorithm}


\subsection{Evaluation of concept coverage metrics}
\label{subsec:evl_concept}
\textbf{Accuracy of VQA in classifying the actions.} We employ a BLIP-based VQA model~\cite{li2022blip} in our study. We randomly sample generated images from SDXL and conduct human evaluation. Experimental results show that VQA achieves an average accuracy of $95.9\%$ and $96.0\%$ in classifying the actions using closed-ended and open-ended questions, respectively. These high accuracies validate the effectiveness and reliability of the proposed metrics.

\textbf{Experiment setup.} We use CLIPScore as a baseline for comparison. To obtain a definitive determination for alignment and misalignment, a fixed threshold $\gamma$ for CLIPScore is required, where a score below $\gamma$ indicates misalignment. Through careful human evaluation, we set $\gamma$ to 0.2. We denote the metric based on CLIPScore as $\text{cov}_{\text{clip}}$. To set up a ground truth, we invite a few volunteers to manually evaluate the generated images in a loose and strict manner, which results in two human evaluation metrics, $\text{human}_{\text{loose}}$ and $\text{human}_{\text{strict}}$, respectively. $\text{human}_{\text{loose}}$ records the percentage of images that capture the desired concept, while $\text{human}_{\text{strict}}$ records the percentage of images that not only capture the desired concept but also exhibit no defects in the generated human. 

We collect 30 human-related concepts, encompassing common actions or interactions of humans with their environments or other objects. Then, for each concept, we generate 500 images featuring this concept and calculate the corresponding values of $\text{cov}_{\text{closed}}$, $\text{cov}_{\text{open}}$, and $\text{cov}_{\text{clip}}$. 

\begin{table}
    \vspace{-4mm}
    \centering
    \caption{Concept coverage results for five human actions with the SDXL model. 
    Results show $\text{cov}_{\text{closed}}$ aligns well with $\text{human}_{\text{loose}}$ and $\text{cov}_{\text{open}}$ align well with $\text{human}_{\text{strict}}$.
    }
    
    \begin{tabular}{c|ccccc}
        \toprule
        Model & run & dance & sing & talk & cry\\ 
        \midrule
        $\text{cov}_{\text{closed}}$ & 98.8$\%$ & 98.2$\%$ & 91.6$\%$ & 19.6$\%$ & 64.0$\%$\\ 
        \midrule
        $\text{cov}_{\text{open}}$ & 98.6$\%$ & 68.8$\%$ & 87.0$\%$ & 14.0$\%$ & 35.6$\%$\\ 
        \midrule
        $\text{cov}_{\text{clip}}$ & 99.0$\%$ & 99.8$\%$ & 96.2$\%$ & 20.4$\%$ & 98.2$\%$\\ 
        \midrule
        $\text{human}_{\text{loose}}$ & 100$\%$ & 99.0$\%$ & 99.0$\%$ & 15.6$\%$ & 71.1$\%$ \\ 
        \midrule
        $\text{human}_{\text{strict}}$ & 91.7$\%$ & 64.2$\%$ & 75.5$\%$ & 11.0$\%$ & 55.0$\%$ \\ 
        \bottomrule
    \end{tabular}
    \label{tab:coverage_analysis_result}
    \vspace{-10pt}
\end{table}

\textbf{Results.} Table \ref{tab:coverage_analysis_result} presents the concept coverage results for five common human action concepts evaluated in the SDXL model. We observe that $\text{cov}_{\text{closed}}$ aligns well with $\text{human}_{\text{loose}}$ and $\text{cov}_{\text{open}}$ align well with $\text{human}_{\text{strict}}$, which are both better than $\text{cov}_{\text{clip}}$. This demonstrates the effectiveness of our proposed metrics. Additional results on more concepts/models are in Appendix 3.1. 


\subsection{Identification of potential bias}
\label{subsec:bias}
In this section, we introduce a VQA-based method to identify potential biases that may exist in recent generative models, with a specific focus on biases related to gender, race, and age. The procedures employed for identifying these biases are similar to those used in concept coverage analysis. Specifically, given a human-related input prompt $\mathbf{p}$, our initial step involves generating a collection of images using this prompt. Then, we apply VQA to extract the gender, race, and age attributes for each generated image. For each attribute, we gather all the corresponding answers and use pre-defined values (gender: \textit{male, female}; race: \textit{White, African, Asian, Indian}; age: \textit{baby, toddler, teenager, middle-aged, old}) to conduct semantic clustering and calculate the semantic entropy. In general, a lower semantic entropy for attribute $i$ indicates a stronger tendency of the model to generate humans with a specific value of attribute $i$ in response to the input prompt $\mathbf{p}$. In other words, it suggests the presence of a bias in the model's generation process for $\mathbf{p}$. 

To automate the bias identification process, we set a threshold for the semantic entropy through careful human evaluation. Specifically, we set $0.8$, $1.0$ and $1.0$ for gender, race, and age, respectively. Notably, the approach can also be used to identify biases in any other human datasets.

\subsection{Evaluation of the bias identification method}
\textbf{Accuracy of VQA.} We gather a substantial dataset of generated images and perform a human evaluation to measure the accuracy of VQA in assessing gender, race, and age attributes. The experimental results demonstrate that VQA achieves an accuracy of $93.80\%$, $92.30\%$, and $84.50\%$ in extracting the gender, race, and age attributes, respectively. The high accuracy attained by VQA ensures the reliability of subsequent fairness analysis.

\textbf{Fairness bias of VQA.} {
The bias of the VQA model may impact the evaluation of the generative model. We analyze the variation of VQA accuracy across different image groups categorized by gender, race, and age. We observe that the accuracy variation is minimal, typically remaining within 1.0\%, indicating that the introduced bias is negligible. Further details are provided in Appendix 4.2.
}

\section{Empirical study on T2I generative models}
\label{sec:experiment}
\subsection{Aesthetic comparison between models}
\label{subsec:aes_gen}
\textbf{Experiment setup.} We compare the aesthetic scores on 50,000 images generated by four popular T2I generative models, including Midjourney, SDXL, SD2.1, and SD1.5. We use diverse prompts adapted from \cite{ju2023humansd} to generate images. Additionally, we refine these prompts to enhance the realism of the generated images. The details of the prompts we use are presented in Appendix 1.3.

\begin{table}
  \centering
  \captionof{table}{Comparison of aesthetic scores for four T2I generative models.}
  \begin{tabular}{c|cccc}
    \toprule
    Model & Midjourney & SDXL & SD2.1 & SD1.5 \\ \midrule
    Mean & 6.35 & 6.23 & 5.62 & 5.54 \\
    Std & 0.45 & 0.55 & 0.57 & 0.55 \\
    \bottomrule
  \end{tabular}
  \label{tab:aesthetic_gen}
  \vspace{-4mm}
\end{table}

\textbf{Results.} The mean and standard deviation of aesthetic scores for the generated images by the models are presented in Table \ref{tab:aesthetic_gen}. Notably, we observe a gradual increase in the average aesthetic scores of the Stable Diffusion (SD) models with each update, while the standard deviation remains unchanged. The images generated by Midjourney exhibit superior aesthetic scores, with a lower standard deviation compared to the SD models. This means that the aesthetic quality of images from Midjourney is more stable. From an aesthetic perspective, we can conclude that SD models are still lagging behind Midjourney.


\subsection{Comparing defect rates between models}
\textbf{Experiment setup.} We assess the defect rates of human faces predicted by our face evaluation model in the images generated by Midjourney, SDXL, SD2.1, and SD1.5. The prompts we use here are the same as in Section \ref{subsec:aes_gen}. Additionally, we present the defect rates of ten human components predicted by our models in Appendix 2.9.

\begin{table}
  \centering
  \captionof{table}{Comparison of defect rates of the human faces in images generated by four T2I generative models.}
  \begin{tabular}{c|cccc}
    \toprule
    Model & Midjourney & SDXL & SD2.1 & SD1.5 \\ \midrule
    Face Defect Rate & 29\% & 61\% & 79\% & 86\% \\
    \bottomrule
  \end{tabular}
  \label{tab:def_gen}
\end{table}
 
\begin{figure}[htbp]
  \centering
   \includegraphics[width=0.75\linewidth]{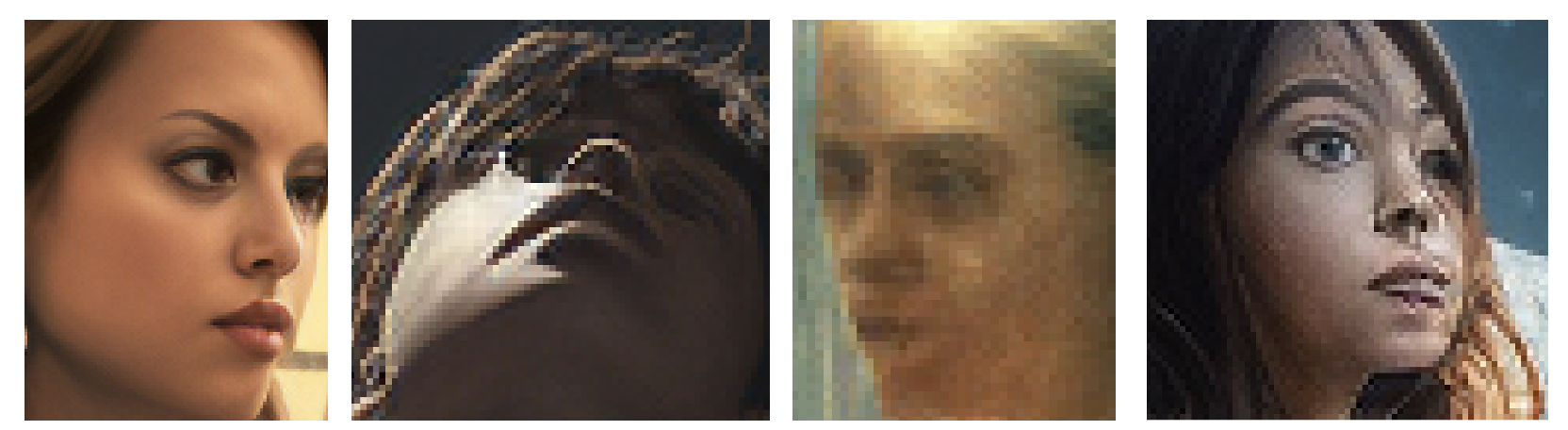}
   \caption{Facial defects from Midjourney, as identified by our face evaluation model.}
   \label{fig:def_mid}
\end{figure}

\textbf{Results.} From Table \ref{tab:def_gen}, we can observe that Midjourney gets the lowest defect rate, demonstrating its superior performance. SDXL also exhibits significant improvement when compared to SD2.1 and SD1.5. Figure \ref{fig:def_mid} presents some defective samples found by our face evaluation model.

\begin{table}
    \centering
    \caption{Concept coverage analysis results for the SDXL, SD2.1, and SD1.5 models. We record the average $\text{cov}_\text{closed}$ and $\text{cov}_\text{open}$ over the 30 concepts for different models. The values with the best performance are highlighted in bold.}
    \begin{tabular}{c|ccc}
        \toprule
        Model & SDXL & SD2.1 & SD1.5 \\ 
        \midrule
        Avg. $\text{cov}_\text{closed}$ & \textbf{94.27$\%$} & 92.93$\%$ & 92.82$\%$\\ 
        \midrule
        Avg. $\text{cov}_\text{open}$ & \textbf{86.50$\%$} & 76.27$\%$ & 77.19$\%$\\ 
        \bottomrule
    \end{tabular}
    \label{tab:coverage_comparison}
    \vspace{-4mm}
\end{table}


\subsection{Concept coverage analysis of generative models}
\textbf{Experiment setup.} We conduct a concept coverage analysis for the SDXL, SD2.1, and SD1.5 models using the same experimental settings as described in Section \ref{subsec:evl_concept}. The models are compared based on the average concept coverage across all 30 collected concepts.

\textbf{Results.} As shown in Table~\ref{tab:coverage_comparison}, SDXL exhibits superior performance in generating images that are coherent with the text inputs, demonstrating its advance in concept understanding. 

\begin{table}
    \centering
    \vspace{-6mm}
    \caption{Fairness analysis results for the gender, race, and age attributes of the SDXL, SD2.1, and SD1.5 models. For each entry, the left value is the proportion of biased prompts and the right value records the average semantic entropy for these biased prompts. The values with the best performance are highlighted in bold.}
    \begin{tabular}{c|ccc}
        \toprule
        Model & SDXL & SD2.1 & SD1.5 \\ 
        \midrule
        Gender Bias & 51$\%$ / 0.45 & \textbf{41$\%$} / \textbf{0.51} & 51$\%$ / 0.46\\ 
        \midrule
        Race Bias & \textbf{27$\%$} / 0.53 & 35$\%$ / \textbf{0.63} & 47$\%$ / 0.54\\ 
        \midrule
        Age Bias & 35$\%$ / 0.59 & 31$\%$ / 0.63 & \textbf{24$\%$} / \textbf{0.65}\\ 
        \bottomrule
    \end{tabular}
    \label{tab:fairness_analysis_result}
    \vspace{-8mm}
\end{table}
\subsection{Fairness analysis of generative models}
\textbf{Experiment setup.} In this section, we conduct a fairness analysis for the SDXL, SD2.1, and SD1.5 models, with a focus on assessing their gender, race, and age biases. Overall, we design a set of 51 human-related prompts to evaluate the models' fairness. For each prompt, we generate 500 images and calculate the semantic entropy for bias identification as described in Section \ref{subsec:bias}. 

We use two metrics to measure the models' tendency to exhibit biases. The first metric measures the proportion of biased prompts, indicating the percentage of prompts where the model exhibits biases. The second metric is the average semantic entropy for these biased prompts. Consequently, a higher proportion of biased prompts and a lower average semantic entropy indicate a stronger tendency of the model to exhibit biased behavior towards a specific attribute value. We present the quantitative results for the fairness analysis in Table \ref{tab:fairness_analysis_result}. Detailed results about prompts and biases are present in Appendix 4.1 \& 4.3. 

\textbf{Results.} Surprisingly, we observe that current generative models have significant fairness issues in human image synthesis. For example, when the input prompt contains information about hair, such as hair color or hair shape, we observe a bias towards generating female images. From SD1.5 to SDXL, despite the overall improvement in the quality of generated images, it is worth noting that fairness is decreasing in many cases. This indicates that image quality and fairness are not always positively related, emphasizing the need to consider fairness in future development. 
\section{Limitations and future works}
\label{sec:limit}

This research represents an initial step toward an intricate understanding of the performance of T2I generative models. We acknowledge that our proposed evaluation framework, while providing substantial insights, does not encompass all aspects of T2I generative model assessment. Ongoing refinement and development of measurement methodologies are imperative to achieve a more exhaustive and nuanced evaluation of these models. For example, while the defect identification model used in our framework shows promise, it also highlights areas for further development and accuracy enhancement. In terms of concept coverage, our current approach is confined to evaluating single concepts per prompt, which cannot completely encapsulate the complexities found in scenarios with multiple interrelated concepts. Future research should aim to broaden the evaluation framework, incorporating a more diverse array of concepts for a richer and more detailed analysis.

\section{Conclusion}
\label{sec:conclude}
In this paper, we have introduced an evaluation framework for T2I generative models, aimed at a balanced examination of image quality and text conditions. Our work, while an initial exploration in this area, contributes to the field with a novel aesthetic score prediction model and a dataset that assists in identifying defects in human image synthesis. Through assessing concept coverage and potential biases in race, gender, and age, our study seeks to provide a more nuanced understanding of these models' capabilities and limitations. The application of our framework across various T2I models, such as SD1.5, SD2.1, SDXL, and Midjourney, has yielded insights that align with human judgments, suggesting its utility in evaluating these systems. We hope our findings will pave the way for future explorations and enhancements in the domain of text-to-image generation.




\newpage
\setcounter{page}{1}
\setcounter{section}{0}

\section{Appendix}
The appendix presents more details and additional results not included in the main paper.

\section{Image aesthetics}
\begin{figure*}[htbp]
  \includegraphics[width=\linewidth]{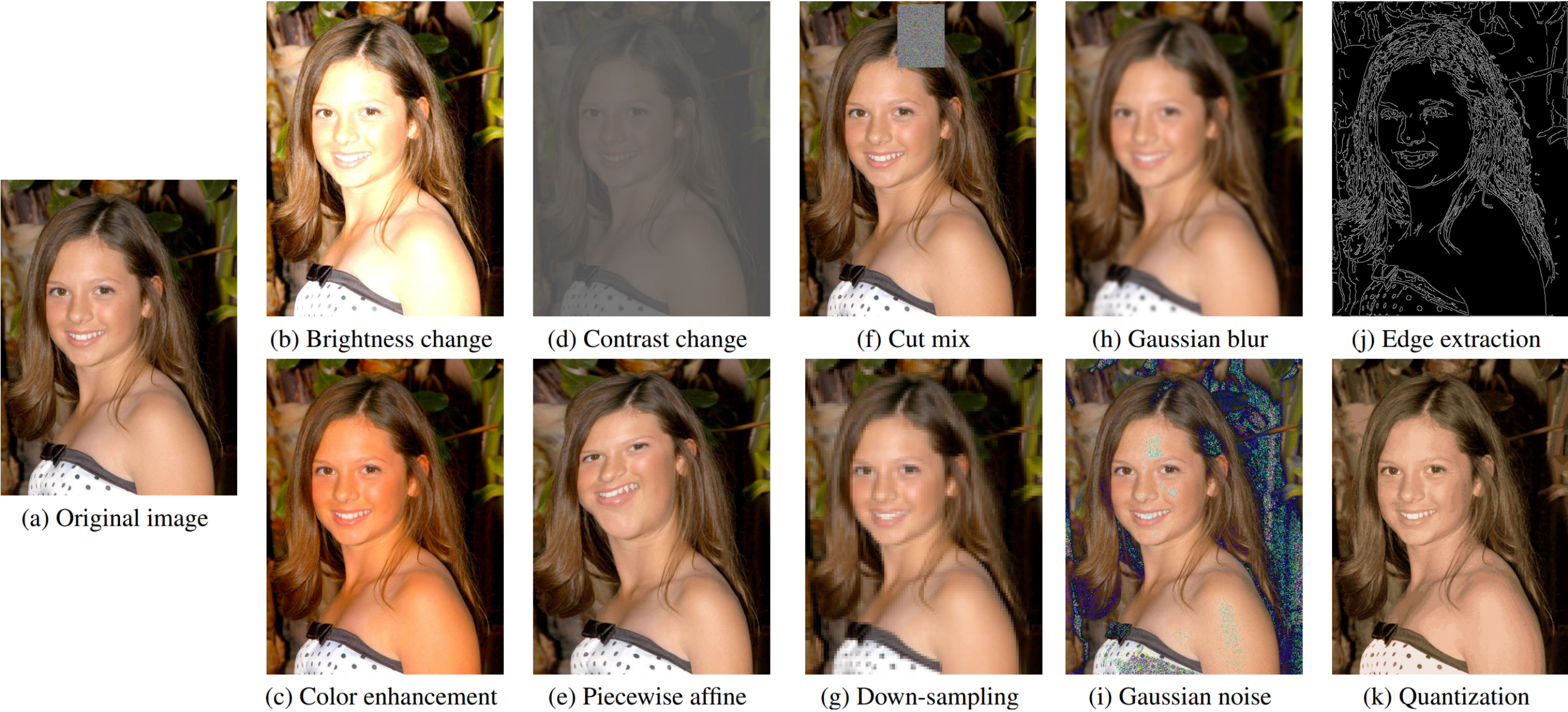}
  \caption{Different distortions applied to an image.}
  \label{distortion_used}
\end{figure*}
\subsection{Distortions used in the distortion prediction task}
During the training process of the generic aesthetic model, we incorporate various distortions to facilitate the module's learning of general aesthetic features, such as the use of color and light, and composition. The distortions applied include \textit{Brightness change, Color enhancement, Contrast change, Piecewise affine, Cut mix, Down-sampling, Gaussian blur, Gaussian noise, Edge extraction (Canny) and Quantization}. Figure~\ref{distortion_used} provides a visualization for these distortions. For detailed implementations, please refer to our code. 

\subsection{Aesthetic attributes prediction} In the main paper, we train CAN on the AVA dataset to predict an overall score for generated images. However, the AVA dataset provides a limited number of training samples for aesthetic attribute prediction. To enhance practical performance, in the released version of CAN, we incorporate additional steps. In addition to pretraining on the AVA dataset, we freeze the entire network and fine-tune only the fully connected layers using the PARA dataset. The PARA dataset offers high-quality attribute annotations, enabling improved aesthetic attribute predictions.

\subsection{Prompts used for the aesthetic comparison}
\begin{figure*}[!t]
    \centering
    \includegraphics[width=0.8\linewidth]{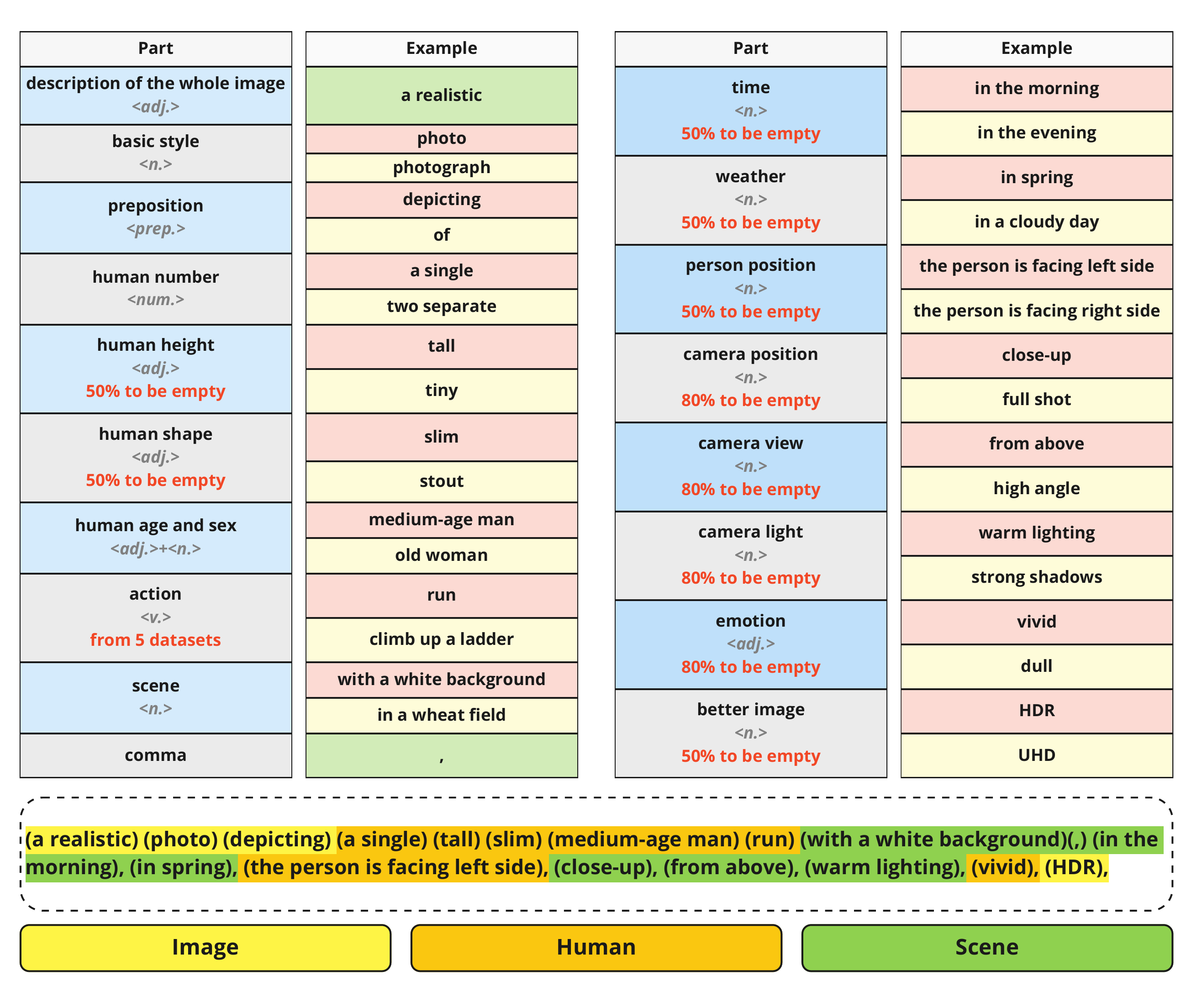}
    \caption{Prompt generation pipeline. Specifically, each prompt is composed of 18 distinct parts to describe three main components: image, human, and scene. And different parts have different probabilities to appear, resulting in diverse and rich prompts.}
    \label{prompt}
\end{figure*}

In the context of aesthetic comparisons, a diverse array of prompts is employed to generate human images for comprehensive assessments. Building upon the generation of human-related prompts outlined in [9], we make modifications to articulate specific image styles and introduce additional constraints, resulting in enhanced realism of the generated images. As illustrated in Figure \ref{prompt}, our refined prompts intricately delineate eighteen distinct components, encompassing descriptions of the entire image (\textit{e.g.}, image style), human features (\textit{e.g.}, human number, height, shape, age, sex, and action), and background scenes (\textit{e.g.}, time, weather, camera settings). This intentional diversification of prompts contributes to the robustness and comprehensiveness of the aesthetic comparison.

\begin{table}[htbp]
  \centering
  \caption{Performance of CAN w/o distortion task or key components on AVA dataset.}
  \resizebox{1.0\textwidth}{!}{
    \setlength\tabcolsep{7.5pt}
    \renewcommand\arraystretch{0.95}
  \begin{tabular}{c|cccc}
    \toprule
    Metric & CAN & w/o distortion & w/o generic & w/o style \\ \midrule
    SRCC & 0.754 & 0.751 & 0.744 & 0.644\\
    Rank Acc. & 0.780 & 0.778 & 0.772 & 0.725\\
    \bottomrule
  \end{tabular}
  }
  \label{tab:aesthetic_abalation}
\end{table}

\subsection{Ablation study of the distortion prediction task and model components}
We conduct an ablation study on the distortion prediction task as well as key components of CAN on AVA dataset. In the case of "w/o distortion", the proposed distortion prediction task is not applied during the training of the generic aesthetic module. Table \ref{tab:aesthetic_abalation} demonstrates the effectiveness of the distortion prediction task and the key components of CAN.

\section{Image realism}
\begin{table}
\vspace{-30pt}
    \centering
    \caption{Human concepts utilized to generate humans with diverse attributes. For Action and Interaction, we use $A_{i}$ and $I_{j}$ to represent the corresponding concepts with indices $i$ and $j$, respectively.}
    \begin{tabular}{c|c}
        \toprule
        Attribute & Values (or Concepts) \\ 
        \midrule
        Action & \makecell[c]{\textit{cry(1), dance(2), eat(3), jump(4), laugh(5),} \\ \textit{run(6), sing(7), sleep(8), talk(9), walk(10)}}\\ 
        \midrule
        Interaction & \makecell[c]{\textit{cooking meal(1), reading book(2),} \\ \textit{taking photo(3), gardening(4), } \\ \textit{painting picture(5), walking dog(6),} \\ \textit{riding bicycle(7), doing yoga(8),} \\ \textit{shopping(9), working on computer(10),} \\ \textit{writing journal(11), playing sports(12),} \\ \textit{exercising(13), playing with children(14),} \\ \textit{hiking(15), playing board games(16),} \\ \textit{doing housework(17), meditating(18),} \\ \textit{attending concert(19), socializing(20)}} \\ 
        \midrule
        Age Stage & \makecell[c]{\textit{baby, toddler, teenager} \\ \textit{middle-aged, elderly}}\\ 
        \midrule
        Ethnicity & \makecell[c]{\textit{African, Asian, White}}\\ 
        \midrule
        Eye Color & \makecell[c]{\textit{blue, brown}}\\ 
        \midrule
        Gender & \makecell[c]{\textit{man, woman}}\\ 
        \midrule
        Hair Color & \makecell[c]{\textit{black, blonde}}\\ 
        \midrule
        Hairstyle & \makecell[c]{\textit{curly, long, short, straight}}\\ 
        \midrule
        Height & \makecell[c]{\textit{short, tall}}\\ 
        \midrule
        Skin Tone & \makecell[c]{\textit{dark skin, fair skin}}\\ 
        \midrule
        Weight & \makecell[c]{\textit{fat, slim}}\\ 
        \bottomrule
    \end{tabular}
    \label{tab:defective_image_dataset}
    \vspace{-30pt}
\end{table}
\subsection{Prompts for generated images in the defective images dataset}
Images without defects are readily available in contemporary real human datasets. Consequently, our defective images dataset primarily focuses on concepts wherein generative models exhibit limitations, particularly in the domain of actions and interactions. Furthermore, we undertake careful consideration of a diverse range of human attributes to ensure a fair distribution. Table \ref{tab:defective_image_dataset} enumerates the conceptual domains addressed during the generation of defective data.

\subsection{Annotations for real images}
For real samples, we use human detection models~\cite{lyu2022rtmdet} to generate face and human body bounding boxes automatically. Next, we assign each of the ten components to a label from the set $\{good, invisible\}$. Since the real samples are collected from existing datasets, we derive their annotations from the original annotations provided in these datasets. For images with key points of the human, we generate the labels based on these key points. For the others, we employ VQA techniques~\cite{li2023blip2}. Specifically, we ask questions like "Is the $Name$ of the person visible in this image?", where $Name$ corresponds to the specific component, such as hand. Finally, we conduct manual refinement to ensure the accuracy and consistency of the annotations.

\subsection{Label distribution of the defective images dataset} 
\begin{figure}
  \centering
   \includegraphics[width=0.7\linewidth]{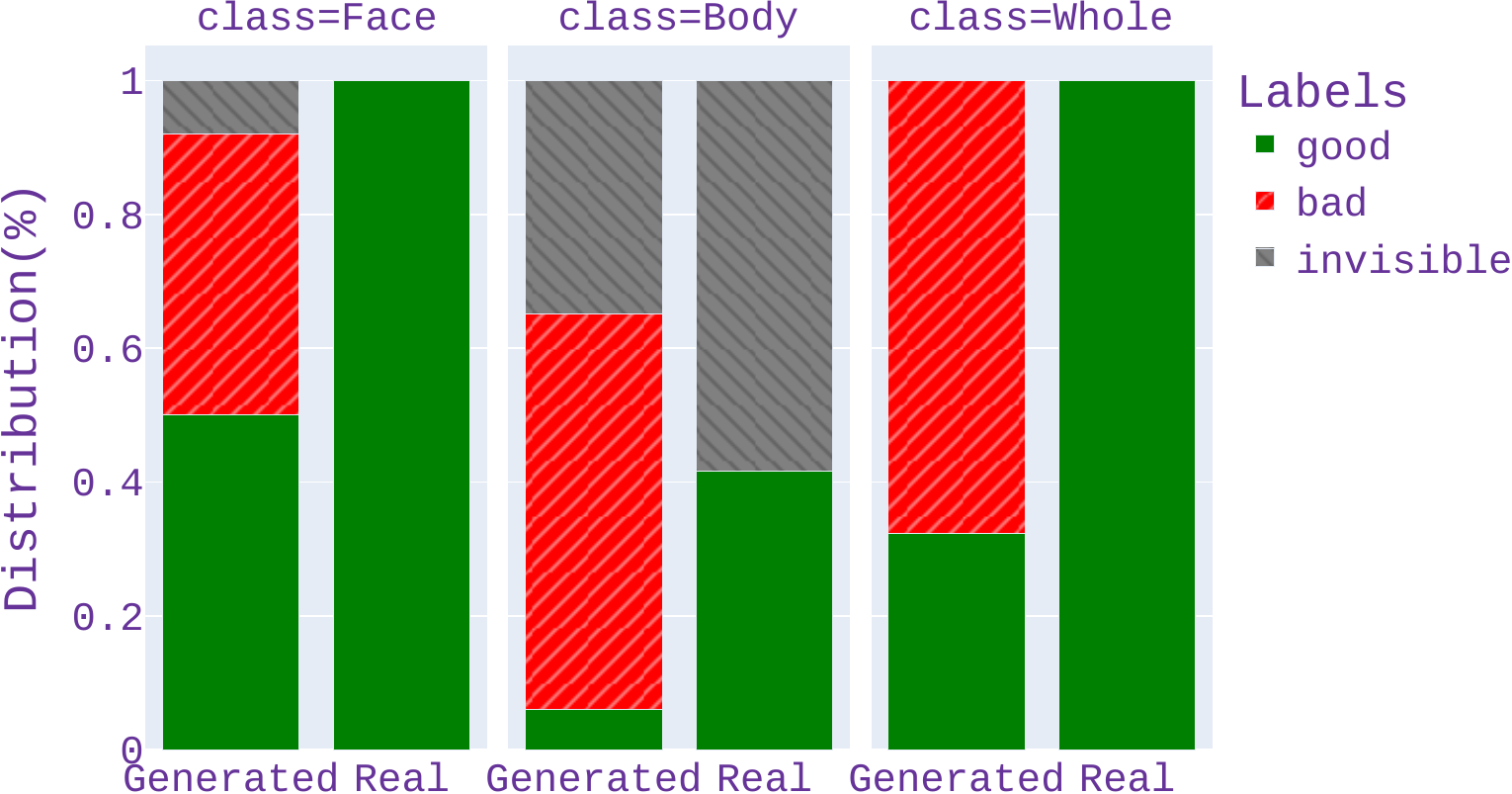}
   \caption[width=\linewidth]{Distribution of coarse-grained labels in the whole dataset.}
   \label{fig:dist_all}
\end{figure}

In Figure \ref{fig:dist_all}, we present the distribution of three coarse-grained labels for both generated and real samples. Due to the high defect rate of generated hands, the label distribution for \textit{body} is imbalanced. However, we can use real samples to rectify this distribution imbalance during training. 

\begin{figure}[htbp]
    \centering
    \includegraphics[width=0.8\linewidth]{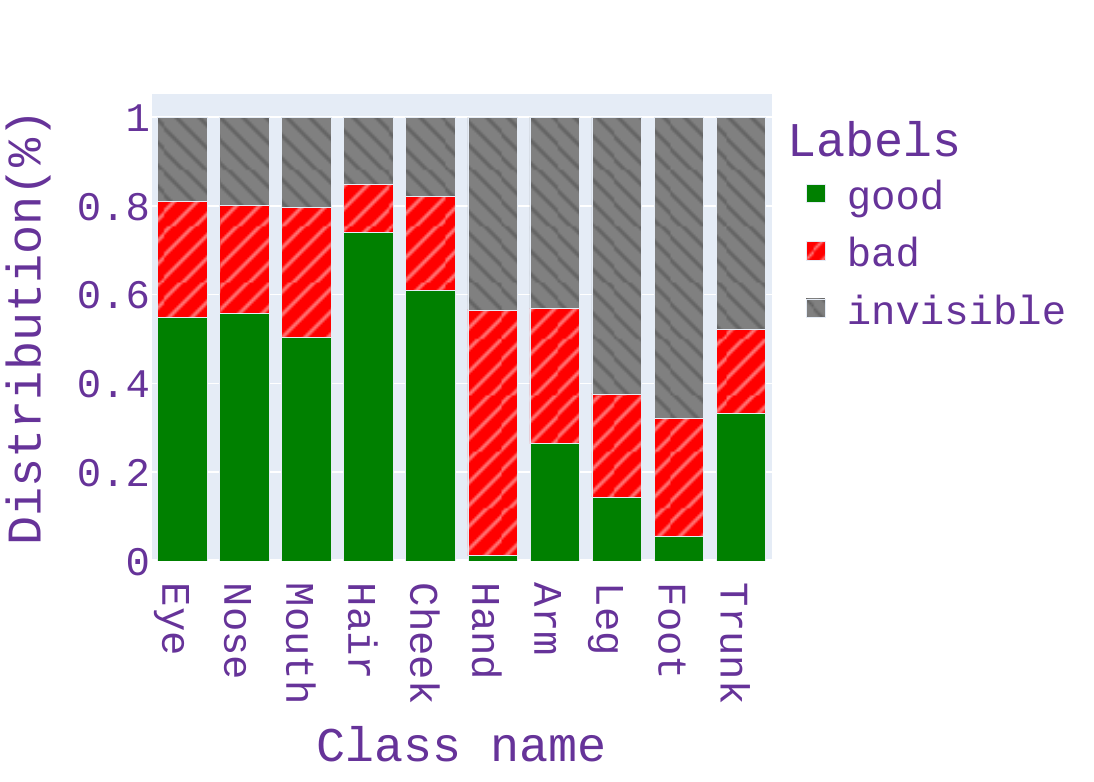}
    \caption{Component-wise label distribution in generated images.}
    \label{fig:neg_dis}
\end{figure}

For individual components. Figure \ref{fig:neg_dis} illustrates a significant observation that highlights a substantial prevalence of defects in the generated hands. This observation underscores the existing limitations in hand generation capabilities, revealing a problem exhibited by SDXL.

\subsection{Can recent metrics identify defects?} IS and FID are commonly used metrics for evaluating the realism of generated images. However, these metrics primarily focus on analyzing image distributions and may not effectively identify defects. Meanwhile, we also find that the subset of images within our dataset that contains defects achieves better FID and IS scores compared to the subset of defect-free images, suggesting that these metrics are insensitive to defects.

\subsection{Can recent pose estimation models and human detection models identify defects?} We explore the potential of leveraging confidence scores obtained from pose estimation models and human detection models to identify defective human components. Our hypothesis is that defective humans or their components would exhibit lower confidence scores in model predictions.

In our experiments, we use RTMDet-l~\cite{lyu2022rtmdet} as the human detection model. The average confidence score for defective generated human images is 0.86, which is lower than 0.92 for defect-free generated human images. However, there is no clear confidence threshold for separating these two sets of images. Besides, the confidence score is affected by the pose and the size of the human in the image.  

For pose estimation models, we use RTMPose-l~\cite{jiang2023rtmpose}. We discover that the confidence scores for output key points associated with defective arms and legs tend to be generally low. However, when it comes to defective hands and face components, the confidence scores remain relatively high. Meanwhile, recent key point detection models rely on low confidence scores to indicate the invisibility of a key point. This characteristic naturally makes it difficult to distinguish between invisible and defective components.

\subsection{Face evaluation model - training details} We use 2k generated images as the test set, while the rest images are used for training. The training process is similar to the classification task, where the cross-entropy loss is applied. 

\subsection{Face evaluation model -  generalization ability} We conduct a human evaluation on the face evaluation model. The accuracies of the model on the hold-out test set of SD1.5, SD2.1, SDXL, and Midjourney are 85.0\%, 84.3\%, 87.2\% and 86.3\%, respectively. These accuracies are close to the 86.0\% accuracy on our validation set, indicating the model's generalization ability and the adequacy of the recent dataset.

\begin{figure*}[h]
    \centering
    \begin{minipage}{0.9\linewidth}
        \subcaptionbox{\textit{false positive}}{
        \includegraphics[width=\linewidth]{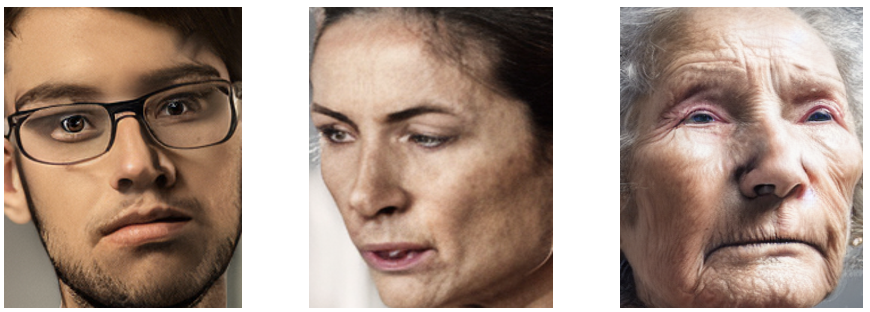}
        }
    \end{minipage}
    \begin{minipage}{0.9\linewidth}
        \subcaptionbox{\textit{false negative}}{
        \includegraphics[width=\linewidth]{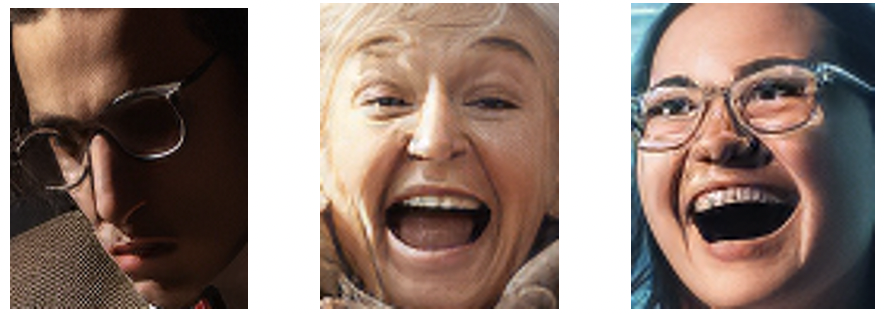}
        }
    \end{minipage}
\caption{We present some hard and typical failure samples in our experiments of face evaluation. We point out that the judgement of these samples exhibits variability among individuals. It is also hard for humans.}
    \label{fig_failure}
\end{figure*}

\subsection{Face evaluation model - failure case analysis}
We conduct a failure case analysis on the face evaluation model. 
We find that typical \textit{False Negative} instances involve images with dramatic facial expressions and loss of facial details due to extreme lighting conditions. Examples can be found in Figure~\ref{fig_failure}(b). On the other hand, typical \textit{False Positive} cases pertain to images with minor defects, particularly abnormalities in eye details, as shown in Figure~\ref{fig_failure}(a).

Importantly, we also recognize that the judgment of slight defects exhibits variability among individuals, introducing inconsistency in the dataset and hindering model improvement. For example, in the middle image of Figure\ref{fig_failure} (a), it is hard to determine whether the distorted eye muscle of the woman is a defect. We are committed to delving further into this nuanced aspect in future studies to better understand and mitigate these challenges.

\subsection{Predicted defect rates on other human components}
\begin{table}
  \centering
  \caption{Comparison of defect rates of ten human components in images generated by SDXL and SD2.1.}
  \begin{tabular}{c|cc}
    \toprule
    Component & SDXL & SD2.1 \\ \midrule
    Eye & 35\%& 50\%\\
    Nose & 42\%& 56\%\\
    Mouth & 48\%& 64\%\\
    Hair & 1\%& 1\%\\
    Cheek & 31\%& 47\%\\
    Hand & 99\%& 98\%\\
    Arm & 38\%& 54\%\\
    Leg & 40\%& 43\%\\
    Foot & 77\%& 78\%\\
    Chunk & 20\%& 37\%\\
    \bottomrule
  \end{tabular}
  \label{tab:def_ten}
  \vspace{-10pt}
\end{table}
In the main paper, we showcase the face defect rates predicted by our face evaluation model. In this section, we present the comparison results predicted by the component-wise evaluation models. As shown in Table~\ref{tab:def_ten}, we can observe that the images generated by SDXL exhibit higher levels of realism compared to those from SD2.1. Moreover, the predicted defect rates of SDXL closely align with the proportion of defective labels in generated images of our defective images dataset, as illustrated in Figure~\ref{fig:neg_dis}. This alignment serves as a testament to the effectiveness of our models in accurately assessing the quality of generated images.

\section{Model concept coverage}
In this section, we provide more detailed results to further validate the effectiveness of the two proposed concept coverage metrics, $\text{cov}_\text{closed}$ and $\text{cov}_\text{open}$. We also provide an explanation on how ChatGPT can be utilized to measure semantic equivalence. Additionally, we provide a visualization that offers a clear view of how to determine whether a generated image accurately represents the desired concept.

\subsection{Detailed results for concept coverage metrics}
For the evaluation of the proposed concept coverage metrics, we adopt 30 human-related concepts, specifically the Action and Interaction concepts listed in Table~\ref{tab:defective_image_dataset}. While the main paper presents results for only five action concepts evaluated in the SDXL model, this section expands upon the results by providing results for all 30 concepts evaluated across the SD1.5, SD2.1, and SDXL models. The concept coverage results for SD1.5, SD2.1, and SDXL are shown in Table~\ref{tab:sd15_coverage_results}
, \ref{tab:sd21_coverage_results}, and \ref{tab:sdxl_coverage_results}, respectively. As observed, in most cases, the results for $\text{cov}_\text{closed}$ correspond well with those of $\text{human}_\text{loose}$, while the results for $\text{cov}_\text{open}$ align well with those of $\text{human}_\text{strict}$. 
Moreover, we calculate the Spearman correlation between human evaluation and the concept coverage metrics using the values in Table~\ref{tab:sd15_coverage_results}
, \ref{tab:sd21_coverage_results}, and \ref{tab:sdxl_coverage_results}. As shown in Table~\ref{tab:spear_cov}, our proposed metrics outperform the baseline method by a significant margin.
These observations validate the effectiveness of our proposed metrics.
\begin{table}[htbp]
  \centering
  \caption{Spearman correlation between human evaluation and the concept coverage metrics. Better results are highlighted in bold.}
  \begin{tabular}{c|ccc}
    \toprule
    Model & SD1.5 & SD2.1 & SDXL \\ \midrule
    $\text{cov}_\text{clip}$-$\text{human}_\text{loose}$ & 0.12 & 0.19 & 0.17\\
    $\text{cov}_\text{closed}$-$\text{human}_\text{loose}$ & \textbf{0.61} & \textbf{0.71} & \textbf{0.48}\\
    $\text{cov}_\text{clip}$-$\text{human}_\text{strict}$ & 0.32 & 0.24 & 0.27\\
    $\text{cov}_\text{open}$-$\text{human}_\text{strict}$ & \textbf{0.51} & \textbf{0.58} & \textbf{0.69}\\
    \bottomrule
  \end{tabular}
  \label{tab:spear_cov}
\end{table}

\subsection{How ChatGPT is used to measure the semantic equivalence?} 
Given two concepts $c_{1}$ and $c_{2}$, we design prompts to query ChatGPT regarding whether one concept could infer the other. If $c_{1}$ could infer $c_{2}$ and vice versa, then $c_{1}$ and $c_{2}$ are deemed semantically equivalent.

\subsection{Visualization for concept coverage analysis}
\begin{figure*}[h]
    \centering
    \begin{minipage}{0.9\linewidth}
        \subcaptionbox{\textit{run}}{
        \includegraphics[width=\linewidth]{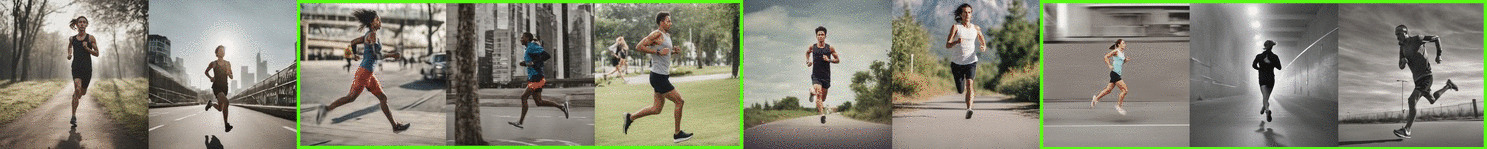}
        }
    \end{minipage}
    \begin{minipage}{0.9\linewidth}
        \subcaptionbox{\textit{jump}}{
        \includegraphics[width=\linewidth]{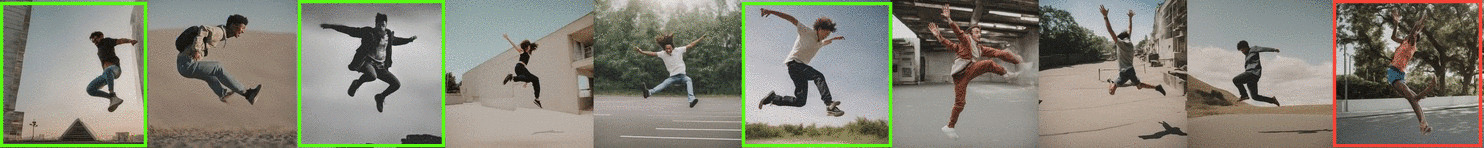}
        }
    \end{minipage}
    \begin{minipage}{0.9\linewidth}
        \subcaptionbox{\textit{dance}}{
        \includegraphics[width=\linewidth]{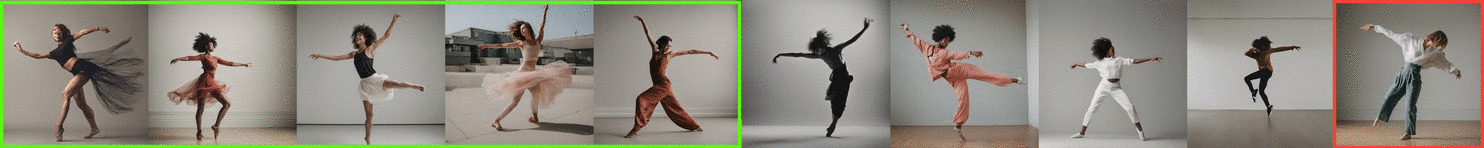}
        }
    \end{minipage}
    \begin{minipage}{0.9\linewidth}
        \subcaptionbox{\textit{sing}}{
        \includegraphics[width=\linewidth]{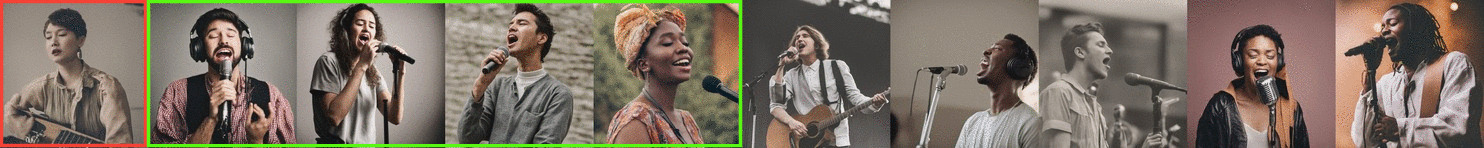}
        }
    \end{minipage}    
    \begin{minipage}{0.9\linewidth}
        \subcaptionbox{\textit{walk}}{
        \includegraphics[width=\linewidth]{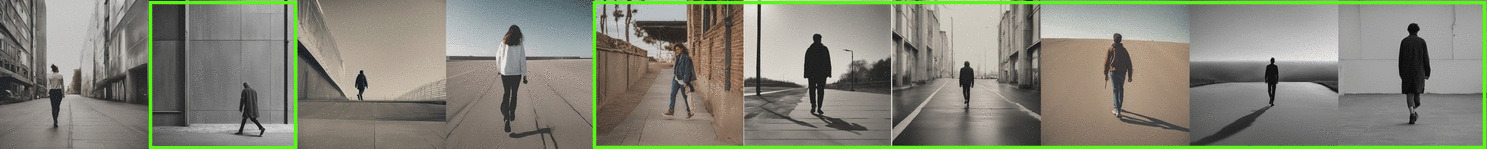}
        }
    \end{minipage}
    \begin{minipage}{0.9\linewidth}
        \subcaptionbox{\textit{talk}}{
        \includegraphics[width=\linewidth]{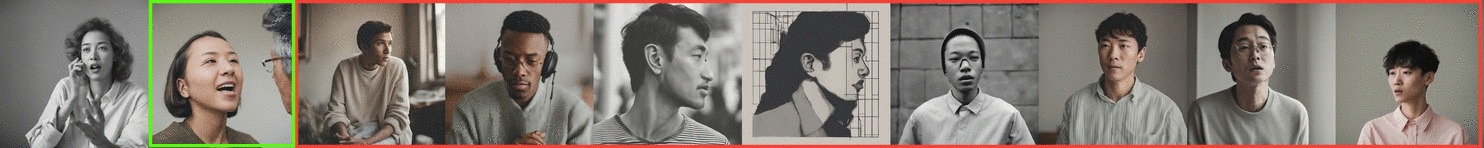}
        }
    \end{minipage}
    \begin{minipage}{0.9\linewidth}
        \subcaptionbox{\textit{sleep}}{
        \includegraphics[width=\linewidth]{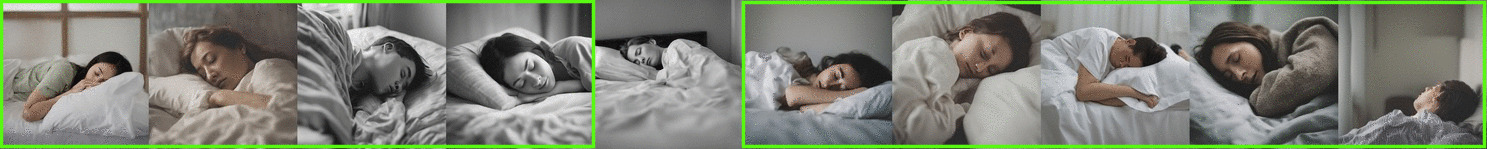}
        }
    \end{minipage}
        \begin{minipage}{0.9\linewidth}
        \subcaptionbox{\textit{eat}}{
        \includegraphics[width=\linewidth]{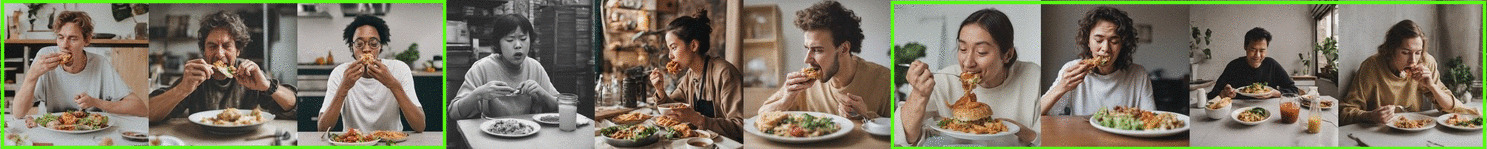}
        }
    \end{minipage}
    \begin{minipage}{0.9\linewidth}
        \subcaptionbox{\textit{cry}}{
        \includegraphics[width=\linewidth]{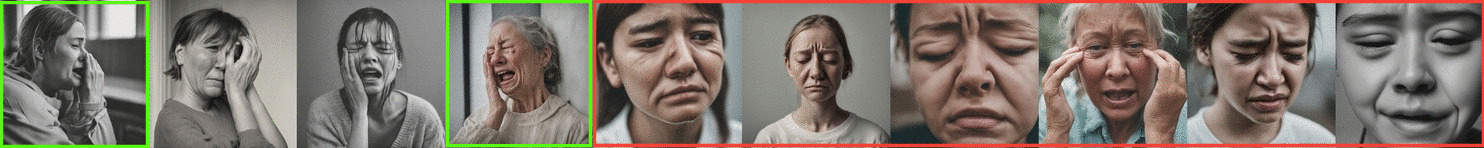}
        }
    \end{minipage}
    \caption{Generated images for ten action concepts in the SDXL model. Images that fail to capture the desired concept are marked with a red bounding box, while images that successfully represent the intended concept with high quality are marked with a green bounding box.}
    \label{fig_coverage_demo}
\end{figure*}

To facilitate a clear understanding of the process for determining whether a generated image accurately represents the intended concept, we provide a visualization in Figure~\ref{fig_coverage_demo}. Specifically, for each Action concept $A_{i}$, we generate 10 images. Images with red bounding boxes are considered ineffective in representing the desired action, while the remaining images (including unmarked ones) successfully depict the correct action (used for calculating $\text{human}_\text{loose}$). And images that successfully represent the intended concept with high quality are marked with a green bounding box (used for calculating $\text{human}_\text{strict}$). We are aware that the annotation process may involve some subjectivity. 
To ensure annotation fairness, annotators are instructed to assign labels based on their initial impressions.

\section{Model fairness}
In this section, we provide the specifics of the 51 human-related prompts used for the fairness analysis. And we provide comprehensive results regarding the fairness bias of VQA. Additionally, we conduct an in-depth analysis of the fairness issues identified in the evaluated models.

\subsection{Prompts for fairness analysis}
We adopt all concepts listed in Table~\ref{tab:defective_image_dataset}, except for $I_{14}$, $I_{19}$, and $I_{20}$. These three concepts involve multiple humans in the generated images, which could potentially impede the fairness analysis. To construct prompts for each concept $c_{i}$, we utilize a simple template, namely "a realistic photo of a person who is $c_{i}$." This approach results in a total of 51 prompts.

\subsection{Detailed results on fairness bias of VQA} 
As our paper concentrates on investigating biases related to gender, race, and age, we conduct comparative experiments to assess the performance of the VQA model across diverse gender, race, and age groups. To achieve this, we initially group the generated images based on their gender, race, and age attributes. Subsequently, we test the accuracies of VQA separately for each group. As shown in Table~\ref{tab:fair_vqa}, the variations in accuracies between different groups is small, indicating the fairness bias introduced by VQA is negligible. However, we note one particular case concerning the accuracy of VQA in extracting the gender attribute from images of Black individuals. In this case, we find it difficult to distinguish between Black males and Black females from the generated images.

\begin{table}[htbp]
  \centering
  \caption{Spearman correlation between human evaluation and the concept coverage metrics. Better results are highlighted in bold.}
  \begin{tabular}{c|ccc}
    \toprule
    Attribute & vqa\_gender & vqa\_race & vqa\_age \\ \midrule
    male & - & $94.29\%$ & $97.14\%$\\
    female & - & $93.82\%$ & $96.30\%$\\ \midrule
    White & $97.18\%$ & - & $95.76\%$\\
    African & $82.61\%$ & - & $95.65\%$\\
    Asian & $96.04\%$ & - & $95.05\%$\\ \midrule
    baby & $96.47\%$ & $91.76\%$ & -\\
    toddler & $96.34\%$ & $93.90\%$ & -\\
    teenager & $98.65\%$ & $95.95\%$ & -\\
    middle-aged & $96.10\%$ & $97.40\%$ & -\\
    old & $100\%$ & $94.52\%$ & -\\
    \bottomrule
  \end{tabular}
  \label{tab:fair_vqa}
\end{table}

\subsection{Detailed fairness analysis results}
Table~\ref{tab:detailed_fairness_analysis} provides a comprehensive summary of the fairness analysis results for the SD1.5, SD2.1, and SDXL models. Specifically, we present the semantic entropies of gender, race, and age attributes for all 51 prompts. Notably, all three models demonstrate significant fairness concerns. For instance, when the input prompts include information about eyes or hair, the models tend to generate female images. Moreover, in the absence of explicit specifications, the models tend to generate humans with White ethnicity.

\onecolumn
{\small
\begin{longtable}{c|ccccc}
    \caption{Concept coverage results for 30 human concepts with the SD1.5 model.}
    \label{tab:sd15_coverage_results}\\
    \toprule
    & & & \textbf{SD1.5} & & \\ 
    Concept & $\text{cov}_\text{closed}$ & $\text{cov}_\text{open}$ & $\text{cov}_\text{clip}$ & $\text{human}_\text{loose}$ & $\text{human}_\text{strict}$ \\ 
    \midrule
    \textit{cry} & 84.0$\%$ &51.6$\%$ & 96.4$\%$ & 78.2$\%$ & 56.2$\%$ \\ 
    \midrule
    \textit{dance} & 94.2$\%$ & 67.0$\%$ & 88.8$\%$ & 93.5$\%$ & 78.3$\%$ \\ 
    \midrule
    \textit{eat} & 98.2$\%$ & 96.6$\%$ & 96.0$\%$ & 93.6$\%$ & 44.0$\%$ \\ 
    \midrule
    \textit{jump} & 74.6$\%$ & 57.4$\%$ & 99.2$\%$ & 95.6$\%$ & 62.6$\%$ \\ 
    \midrule
    \textit{laugh} & 99.4$\%$ & 59.8$\%$ & 98.0$\%$ & 92.5$\%$ & 64.7$\%$ \\ 
    \midrule
    \textit{run} & 99.2$\%$ &  98.4$\%$ & 97.2$\%$ & 99.3$\%$ & 84.2$\%$ \\
    \midrule
    \textit{sing} & 96.2$\%$ & 92.8$\%$ & 98.2$\%$ & 94.5$\%$ & 73.6$\%$ \\ 
    \midrule
    \textit{sleep} & 97.0$\%$ & 97.6$\%$ & 97.2$\%$ & 97.7$\%$ & 63.6$\%$ \\ 
    \midrule
    \textit{talk} & 43.8$\%$ & 33.0$\%$ & 22.4$\%$ & 48.7$\%$ & 27.0$\%$ \\ 
    \midrule
    \textit{walk} & 96.0$\%$ & 35.0$\%$ & 97.6$\%$ & 97.7$\%$ & 55.8$\%$ \\ 
    \midrule
    \textit{cooking meal} & 99.6$\%$ & 99.0$\%$ & 100$\%$ & 98.8$\%$ & 85.3$\%$ \\ 
    \midrule
    \textit{reading book} & 96.4$\%$ & 83.0$\%$ & 89.0$\%$ & 90.3$\%$ & 43.3$\%$ \\ 
    \midrule
    \textit{taking photo} & 99.2$\%$ & 92.4$\%$ & 100$\%$ & 96.6$\%$ & 80.1$\%$ \\ 
    \midrule
    \textit{gardening} & 100$\%$ & 97.0$\%$ & 97.6$\%$ & 97.7$\%$ & 68.3$\%$ \\ 
    \midrule
    \textit{painting picture} & 95.8$\%$ & 98.6$\%$ & 99.6$\%$ & 92.1$\%$ & 87.8$\%$ \\ 
    \midrule
    \textit{walking dog} & 97.8$\%$ & 35.0$\%$ & 99.4$\%$ & 92.7$\%$ & 30.9$\%$ \\ 
    \midrule
    \textit{riding bicycle} & 99.6$\%$ & 82.6$\%$ & 90.0$\%$ & 100$\%$ & 71.4$\%$ \\ 
    \midrule
    \textit{doing yoga} & 94.0$\%$ & 31.2$\%$ & 97.6$\%$ & 98.4$\%$ & 20.7$\%$ \\ 
    \midrule
    \textit{shopping} & 97.8$\%$ & 34.8$\%$ & 99.4$\%$ & 100$\%$ & 25.5$\%$ \\ 
    \midrule
    \textit{working on computer} & 97.6$\%$ & 91.0$\%$ & 99.8$\%$ & 91.0$\%$ & 83.5$\%$ \\ 
    \midrule
    \textit{writing journal} & 98.2$\%$ & 97.2$\%$ & 99.2$\%$ & 87.7$\%$ & 75.3$\%$ \\ 
    \midrule
    \textit{playing sports} & 96.8$\%$ & 96.0$\%$ & 99.8$\%$ & 100$\%$ & 93.4$\%$ \\ 
    \midrule
    \textit{exercising} & 98.6$\%$ & 24.8$\%$ & 99.8$\%$ & 99.4$\%$ & 27.3$\%$ \\ 
    \midrule
    \textit{playing with children} & 100$\%$ & 68.2$\%$ & 77.0$\%$ & 100$\%$ & 43.7$\%$ \\ 
    \midrule
    \textit{hiking} & 99.4$\%$ & 77.4$\%$ & 98.6$\%$ & 100$\%$ & 62.9$\%$ \\ 
    \midrule
    \textit{playing board games} & 96.4$\%$ & 97.2$\%$ & 95.0$\%$ & 97.7$\%$ & 41.4$\%$ \\ 
    \midrule
    \textit{doing housework} & 95.6$\%$ & 99.4$\%$ & 90.4$\%$ & 96.3$\%$ & 38.4$\%$ \\ 
    \midrule
    \textit{meditating} & 39.4$\%$ & 97.0$\%$ & 99.8$\%$ & 68.4$\%$ & 42.8$\%$ \\ 
    \midrule
    \textit{attending concert} & 99.8$\%$ & 96.0$\%$ & 100$\%$ & 100$\%$ & 68.4$\%$ \\ 
    \midrule
    \textit{socializing} & 100$\%$ & 88.6$\%$ & 97.6$\%$ & 100$\%$ & 84.4$\%$ \\ 
    \bottomrule
\end{longtable}
}
\twocolumn

\onecolumn
{\small
\begin{longtable}{c|ccccc}
    \caption{Concept coverage results for 30 human concepts with the SD2.1 model.}
    \label{tab:sd21_coverage_results}\\
    \toprule
    & & & \textbf{SD2.1} & & \\ 
    Concept & $\text{cov}_\text{closed}$ & $\text{cov}_\text{open}$ & $\text{cov}_\text{clip}$ & $\text{human}_\text{loose}$ & $\text{human}_\text{strict}$ \\ 
    \midrule
    \textit{cry} & 73.4$\%$ & 24.8$\%$ & 96.4$\%$ & 75.7$\%$ & 28.7$\%$ \\ 
    \midrule
    \textit{dance} & 92.8$\%$ & 38.4$\%$ & 99.2$\%$ & 88.3$\%$ & 69.6$\%$ \\ 
    \midrule
    \textit{eat} & 99.0$\%$ & 98.2$\%$ & 99.4$\%$ & 99.5$\%$ & 45.8$\%$ \\ 
    \midrule
    \textit{jump} & 69.4$\%$ & 61.0$\%$ & 100$\%$ & 76.6$\%$ & 64.3$\%$ \\ 
    \midrule
    \textit{laugh} & 97.2$\%$ & 63.6$\%$ & 100$\%$ & 84.9$\%$ & 77.5$\%$ \\ 
    \midrule
    \textit{run} & 99.4$\%$ & 96.4$\%$ & 99.6$\%$ & 100$\%$ & 92.4$\%$ \\ 
    \midrule
    \textit{sing} & 98.4$\%$ & 97.0$\%$ & 99.8$\%$ & 92.5$\%$ & 79.4$\%$ \\ 
    \midrule
    \textit{sleep} & 97.4$\%$ & 91.8$\%$ & 97.0$\%$ & 96.0$\%$ & 87.7$\%$ \\ 
    \midrule
    \textit{talk} & 27.4$\%$ & 14.0$\%$ & 39.6$\%$ & 35.8$\%$ & 24.0$\%$ \\ 
    \midrule
    \textit{walk} & 94.0$\%$ & 66.8$\%$ & 99.6$\%$ & 74.4$\%$ & 63.6$\%$ \\ 
    \midrule
    \textit{cooking meal} & 98.6$\%$ & 96.8$\%$ & 100$\%$ & 93.6$\%$ & 47.6$\%$ \\ 
    \midrule
    \textit{reading book} & 98.8$\%$ & 48.2$\%$ & 97.0$\%$ & 92.6$\%$ & 44.6$\%$ \\ 
    \midrule
    \textit{taking photo} & 98.8$\%$ & 95.6$\%$ & 100$\%$ & 95.0$\%$ & 73.3$\%$ \\ 
    \midrule
    \textit{gardening} & 99.6$\%$ & 92.6$\%$ & 99.0$\%$ & 96.5$\%$ & 65.7$\%$ \\ 
    \midrule
    \textit{painting picture} & 98.4$\%$ & 79.4$\%$ & 97.2$\%$ & 96.5$\%$ & 65.8$\%$ \\ 
    \midrule
    \textit{walking dog} & 95.6$\%$ & 66.6$\%$ & 97.8$\%$ & 83.3$\%$ & 58.3$\%$ \\ 
    \midrule
    \textit{riding bicycle} & 99.2$\%$ & 36.0$\%$ & 99.8$\%$ & 100$\%$ & 23.7$\%$ \\ 
    \midrule
    \textit{doing yoga} & 96.2$\%$ & 27.0$\%$ & 99.2$\%$ & 100$\%$ & 16.7$\%$ \\ 
    \midrule
    \textit{shopping} & 99.4$\%$ & 91.6$\%$ & 100$\%$ & 100$\%$ & 53.3$\%$ \\ 
    \midrule
    \textit{working on computer} & 100$\%$ & 82.2$\%$ & 99.4$\%$ & 100$\%$ & 75.1$\%$ \\ 
    \midrule
    \textit{writing journal} & 98.4$\%$ & 98.4$\%$ & 100$\%$ & 98.3$\%$ & 80.2$\%$ \\ 
    \midrule
    \textit{playing sports} & 85.0$\%$ & 78.8$\%$ & 100$\%$ & 96.7$\%$ & 86.4$\%$ \\ 
    \midrule
    \textit{exercising} & 97.4$\%$ & 97.2$\%$ & 100$\%$ & 85.7$\%$ & 64.7$\%$ \\ 
    \midrule
    \textit{playing with children} & 99.2$\%$ & 97.8$\%$ & 98.2$\%$ & 100$\%$ & 78.0$\%$ \\ 
    \midrule
    \textit{hiking} & 99.4$\%$ & 97.8$\%$ & 100$\%$ & 100$\%$ & 84.8$\%$ \\ 
    \midrule
    \textit{playing board games} & 96.8$\%$ & 98.0$\%$ & 100$\%$ & 100$\%$ & 76.7$\%$ \\ 
    \midrule
    \textit{doing housework} & 94.4$\%$ & 96.2$\%$ & 98.2$\%$ & 96.7$\%$ & 86.7$\%$ \\ 
    \midrule
    \textit{meditating} & 85.0$\%$ & 78.8$\%$ & 99.6$\%$ & 87.0$\%$ & 71.7$\%$ \\ 
    \midrule
    \textit{attending concert} & 99.4$\%$ & 85.0$\%$ & 99.6$\%$ & 100$\%$ & 58.3$\%$ \\ 
    \midrule
    \textit{socializing} & 99.8$\%$ & 38.6$\%$ & 99.6$\%$ & 100$\%$ & 35.0$\%$ \\ 
    \bottomrule
\end{longtable}
}
\twocolumn

\onecolumn
{\small
\begin{longtable}{c|ccccc}
    \caption{Concept coverage results for 30 human concepts with the SDXL model.}
    \label{tab:sdxl_coverage_results}\\
    \toprule
    & & & \textbf{SDXL} & & \\ 
    Concept & $\text{cov}_\text{closed}$ & $\text{cov}_\text{open}$ & $\text{cov}_\text{clip}$ & $\text{human}_\text{loose}$ & $\text{human}_\text{strict}$ \\ 
    \midrule
    \textit{cry} & 64.0$\%$ & 35.6$\%$ & 98.2$\%$ & 71.1$\%$ & 55.0$\%$ \\ 
    \midrule
    \textit{dance} & 98.2$\%$ & 68.8$\%$ & 99.8$\%$ & 99.0$\%$ & 64.2$\%$ \\ 
    \midrule
    \textit{eat} & 98.8$\%$ & 98.8$\%$ & 100$\%$ & 96.9$\%$ & 82.1$\%$ \\ 
    \midrule
    \textit{jump} & 93.4$\%$ & 78.8$\%$ & 100$\%$ & 100$\%$ & 59.4$\%$ \\ 
    \midrule
    \textit{laugh} & 99.6$\%$ & 81.2$\%$ & 100$\%$ & 96.8$\%$ & 85.3$\%$ \\ 
    \midrule
    \textit{run} & 98.8$\%$ & 98.6$\%$ & 99.0$\%$ & 100$\%$ & 91.7$\%$ \\ 
    \midrule
    \textit{sing} & 91.6$\%$ & 87.0$\%$ & 96.2$\%$ & 99.0$\%$ & 75.5$\%$ \\ 
    \midrule
    \textit{sleep} & 96.8$\%$ & 87.6$\%$ & 98.8$\%$ & 98.0$\%$ & 52.9$\%$ \\ 
    \midrule
    \textit{talk} & 19.6$\%$ & 14.0$\%$ & 20.4$\%$ & 15.6$\%$ & 11.0$\%$ \\ 
    \midrule
    \textit{walk} & 98.8$\%$ & 96.8$\%$ & 99.8$\%$ & 100$\%$ & 88.1$\%$ \\ 
    \midrule
    \textit{cooking meal} & 99.4$\%$ & 99.4$\%$ & 100$\%$ & 97.1$\%$ & 78.8$\%$ \\ 
    \midrule
    \textit{reading book} & 94.6$\%$ & 94.0$\%$ & 97.6$\%$ & 97.9$\%$ & 68.7$\%$ \\ 
    \midrule
    \textit{taking photo} & 99.8$\%$ & 95.8$\%$ & 99.2$\%$ & 100$\%$ & 82.0$\%$ \\ 
    \midrule
    \textit{gardening} & 100$\%$ & 96.6$\%$ & 100$\%$ & 100$\%$ & 72.4$\%$ \\ 
    \midrule
    \textit{painting picture} & 99.6$\%$ & 98.8$\%$ & 100$\%$ & 100$\%$ & 63.4$\%$ \\ 
    \midrule
    \textit{walking dog} & 99.6$\%$ & 99.4$\%$ & 99.8$\%$ & 98.8$\%$ & 91.8$\%$ \\ 
    \midrule
    \textit{riding bicycle} & 100$\%$ & 99.2$\%$ & 100$\%$ & 87.9$\%$ & 74.4$\%$ \\ 
    \midrule
    \textit{doing yoga} & 99.4$\%$ & 90.2$\%$ & 100$\%$ & 100$\%$ & 75.1$\%$ \\ 
    \midrule
    \textit{shopping} & 100$\%$ & 99.0$\%$ & 100$\%$ & 100$\%$ & 84.9$\%$ \\ 
    \midrule
    \textit{working on computer} & 91.6$\%$ & 91.0$\%$ & 99.2$\%$ & 90.6$\%$ & 44.1$\%$ \\ 
    \midrule
    \textit{writing journal} & 99.2$\%$ & 98.8$\%$ & 99.6$\%$ & 100$\%$ & 85.0$\%$ \\ 
    \midrule
    \textit{playing sports} & 99.8$\%$ & 79.4$\%$ & 97.0$\%$ & 100$\%$ & 71.0$\%$ \\ 
    \midrule
    \textit{exercising} & 99.4$\%$ & 65.8$\%$ & 99.0$\%$ & 91.2$\%$ & 42.3$\%$ \\ 
    \midrule
    \textit{playing with children} & 99.8$\%$ & 45.8$\%$ & 98.8$\%$ & 100$\%$ & 21.2$\%$ \\ 
    \midrule
    \textit{hiking} & 100$\%$ & 98.2$\%$ & 99.6$\%$ & 100$\%$ & 85.0$\%$ \\ 
    \midrule
    \textit{playing board games} & 94.2$\%$ & 93.2$\%$ & 99.6$\%$ & 91.0$\%$ & 85.5$\%$ \\ 
    \midrule
    \textit{doing housework} & 97.2$\%$ & 88.2$\%$ & 98.4$\%$ & 100$\%$ & 63.6$\%$ \\ 
    \midrule
    \textit{meditating} & 95.8$\%$ & 91.4$\%$ & 99.8$\%$ & 95$\%$ & 87.6$\%$ \\ 
    \midrule
    \textit{attending concert} & 99.0$\%$ & 33.8$\%$ & 99.8$\%$ & 96.6$\%$ & 52.8$\%$ \\ 
    \midrule
    \textit{socializing} & 100$\%$ & 49.8$\%$ & 100$\%$ & 100$\%$ & 43.7$\%$ \\ 
    \bottomrule
\end{longtable}
}
\twocolumn

\onecolumn
{\small
\begin{longtable}{c|ccc|ccc|ccc}
    \caption{The semantic entropies of gender, race, and age attributes for 51 concepts (or prompts) evaluated in the SDXL, SD2.1, and SD1.5 models. Bold symbols indicate instances where models exhibit biases.}
    \label{tab:detailed_fairness_analysis}\\
    \toprule
     & & \textbf{SDXL} & & & \textbf{SD2.1} & & & \textbf{SD1.5} & \\ 
    Concept & Gender & Race & Age & Gender & Race & Age & Gender & Race & Age \\ 
    \midrule
    \endfirsthead
    \multicolumn{10}{r}{Continued}\\
    \toprule
     & & \textbf{SDXL} & & & \textbf{SD2.1} & & & \textbf{SD1.5} & \\ 
    Concept & Gender & Race & Age & Gender & Race & Age & Gender & Race & Age \\ 
    \midrule
    \endhead
    \textit{cry} & \textbf{0.43} & \textbf{0.91} & 1.79 & 1.00 & 1.47 & 1.36 & \textbf{0.33} & 1.05 & 1.72 \\ 
    \midrule
    \textit{dance} & 0.85 & 1.38 & 1.32 & 0.99 & 1.16 & 1.31 & 0.87 & 1.45 & 1.21 \\ 
    \midrule
    \textit{eat} & 0.91 & 1.60 & 1.53 & 0.95 & \textbf{0.94} & 1.12 & 0.94 & 1.19 & 1.50 \\ 
    \midrule
    \textit{jump} & \textbf{0.30} & 1.16 & 1.21 & \textbf{0.70} & 1.16 & 1.16 & 1.00  & \textbf{0.46} & 1.12 \\ 
    \midrule
    \textit{laugh} & 0.82 & 1.51 & 1.60 & 0.93 & \textbf{0.92} & 1.16 & 0.93 & 1.50 & 1.64 \\ 
    \midrule
    \textit{run} & 0.99 & 1.36 & 1.02 & 0.94 & 1.10 & 1.22 & 1.00 & \textbf{0.88} & 1.14 \\ 
    \midrule
    \textit{sing} & 0.85 & 1.40 & 1.41 & \textbf{0.57} & 1.17 & 1.27 & \textbf{0.62} & 1.62 & 1.23 \\ 
    \midrule
    \textit{sleep} & 0.92 & 1.27 & 1.28 & 1.00 & 1.12 & 1.46 & 0.97 & 1.40 & 1.40 \\ 
    \midrule
    \textit{talk} & 1.00 & 1.57 & 1.60 & 0.95 & 1.15 & \textbf{0.60} & 0.98 & 1.45 & 1.31 \\ 
    \midrule
    \textit{walk} & \textbf{0.61} & 1.15 & 1.23 & 0.99 & \textbf{0.87} & 1.72 & 0.91 & \textbf{0.35} & 1.62 \\ 
    \midrule
    \textit{cooking meal} & 0.93 & 1.64 & 1.23 & 0.98 & 1.01 & \textbf{0.80} & 0.84 & 1.24 & \textbf{0.97} \\ 
    \midrule
    \textit{reading book} & 1.00 & 1.42 & 1.57 & \textbf{0.56} & 1.54 & 1.42 & 0.84 & 1.42 & 1.54 \\ 
    \midrule
    \textit{taking photo} & 0.96 & 1.17 & 1.61 & 0.98 & 1.02 & 1.48 & 1.00 & \textbf{0.89} & 1.33 \\ 
    \midrule
    \textit{gardening} & 0.91 & 1.32 & 1.37 & \textbf{0.46} & \textbf{0.77} & 1.21 & \textbf{0.77} & \textbf{0.94} & 1.44 \\ 
    \midrule
    \textit{painting picture} & 0.80 & 1.27 & 1.29 & 0.99 & 1.05 & \textbf{0.80} & \textbf{0.72} & \textbf{0.79} & 1.38 \\ 
    \midrule
    \textit{walking dog} & 1.00 & \textbf{0.12} & \textbf{0.31} & 0.95 & \textbf{0.20} & \textbf{0.65} & \textbf{0.60} & \textbf{0.12} & \textbf{0.37} \\ 
    \midrule
    \textit{riding bicycle} & \textbf{0.73} & 1.32 & \textbf{0.83} & 0.98 & 1.28 & 1.19 & \textbf{0.78} & 1.11 & 1.12 \\ 
    \midrule
    \textit{doing yoga} & \textbf{0.31} & 1.35 & 1.60 & \textbf{0.55} & 1.39 & 1.34 & \textbf{0.15} & \textbf{0.77} & 1.03 \\ 
    \midrule
    \textit{shopping} & 0.97 & 1.46 & 1.19 & \textbf{0.36} & 1.15 & 1.18 & \textbf{0.70} & 1.22 & 1.29 \\ 
    \midrule
    \textit{working on computer} & \textbf{0.63} & 1.57 & 1.13 & \textbf{0.67} & 1.77 & \textbf{0.93} & \textbf{0.69} & 1.14 & 1.16 \\ 
    \midrule
    \textit{writing journal} & 0.98 & \textbf{0.86} & \textbf{0.78} & 0.94 & \textbf{0.34} & \textbf{0.55} & \textbf{0.75} & \textbf{0.82} & 1.15 \\ 
    \midrule
    \textit{playing sports} & 0.94 & 1.44 & \textbf{0.91} & \textbf{0.67} & 1.36 & 1.39 & 0.92 & 1.37 & 1.20 \\ 
    \midrule
    \textit{exercising} & \textbf{0.54} & 1.44 & \textbf{0.96} & \textbf{0.53} & 1.39 & 1.23 & \textbf{0.18} & \textbf{0.80} & \textbf{0.75} \\ 
    \midrule
    \textit{hiking} & 0.89 & \textbf{0.13} & \textbf{0.47} & 0.91 & \textbf{0.39} & \textbf{0.52} & 0.98 & \textbf{0.19} & \textbf{0.50} \\ 
    \midrule
    \textit{playing board games} & \textbf{0.77} & \textbf{0.79} & 1.15 & 0.88 & \textbf{0.79} & 1.36 & 0.99 & \textbf{0.38} & 1.44 \\ 
    \midrule
    \textit{doing housework} & \textbf{0.29} & 1.49 & 1.08 & \textbf{0.49} & 1.42 & 1.06 & \textbf{0.30} & 1.30 & 1.23 \\ 
    \midrule
    \textit{meditating} & 0.98 & 1.33 & 1.64 & 0.84 & 1.73 & 1.33 & 0.94 & 1.66 & 1.37 \\ 
    \midrule
    \textit{baby} & \textbf{0.76} & \textbf{0.94} & \textbf{0.00} & \textbf{0.67} & \textbf{0.98} & \textbf{0.00} & 0.95 & \textbf{0.72} & \textbf{0.00} \\ 
    \midrule
    \textit{toddler} & 0.92 & 1.36 & \textbf{0.30} & 0.94 & 1.08 & \textbf{0.08} & 1.00 & \textbf{0.92} & \textbf{0.02} \\ 
    \midrule
    \textit{teenager} & 0.87 & 1.23 & \textbf{0.97} & \textbf{0.51} & \textbf{0.80} & \textbf{0.96} & 0.96 & 1.02 & \textbf{0.43} \\ 
    \midrule
    \textit{middle-aged} & \textbf{0.69} & \textbf{0.21} & \textbf{0.78} & 0.83 & 1.01 & \textbf{0.91} & \textbf{0.41} & \textbf{0.05} & \textbf{0.84} \\ 
    \midrule
    \textit{elderly} & \textbf{0.67} & \textbf{0.14} & \textbf{0.04} & \textbf{0.47} & \textbf{0.09} & \textbf{0.05} & 0.91 & \textbf{0.14} & \textbf{0.00} \\ 
    \midrule
    \textit{African} & 0.92 & \textbf{0.00} & 1.18 & 0.98 & \textbf{0.00} & 1.53 & 0.89 & \textbf{0.00} & 1.26 \\ 
    \midrule
    \textit{Asian} & 0.92 & \textbf{0.00} & 1.75 & 0.96 & \textbf{0.02} & 1.34 & \textbf{0.69} & \textbf{0.00} & 1.40 \\ 
    \midrule
    \textit{White} & \textbf{0.46} & \textbf{0.49} & 1.37 & \textbf{0.28} & \textbf{0.62} & 1.15 & 0.98 & 1.00 & 1.10 \\ 
    \midrule
    \textit{blue eye} & \textbf{0.34} & \textbf{0.02} & \textbf{0.71} & \textbf{0.61} & \textbf{0.20} & \textbf{0.93} & \textbf{0.48} & \textbf{0.02} & \textbf{0.95} \\ 
    \midrule
    \textit{brown eye} & \textbf{0.24} & \textbf{0.50} & \textbf{0.45} & 0.87 & 1.50 & 1.28 & \textbf{0.33} & 1.03 & 1.03 \\ 
    \midrule
    \textit{man} & \textbf{0.05} & 1.04 & \textbf{0.74} & \textbf{0.02} & \textbf{0.74} & \textbf{0.22} & \textbf{0.00} & \textbf{0.51} & \textbf{0.36} \\ 
    \midrule
    \textit{woman} & \textbf{0.00} & \textbf{0.92} & \textbf{0.82} & \textbf{0.04} & 1.06 & \textbf{0.50} & \textbf{0.04} & \textbf{0.83} & 1.22 \\ 
    \midrule
    \textit{black hair} & \textbf{0.66} & \textbf{0.32} & \textbf{0.62} & \textbf{0.63} & 1.36 & 1.00 & \textbf{0.19} & 1.58 & \textbf{0.77} \\ 
    \midrule
    \textit{blonde hair} & \textbf{0.02} & \textbf{0.00} & \textbf{0.19} & \textbf{0.29} & \textbf{0.00} & \textbf{0.53} & \textbf{0.05} & \textbf{0.04} & \textbf{0.62} \\ 
    \midrule
    \textit{curly hair} & \textbf{0.43} & \textbf{0.32} & \textbf{0.98} & \textbf{0.65} & \textbf{0.90} & 1.25 & \textbf{0.34} & \textbf{0.24} & 1.12 \\ 
    \midrule
    \textit{long hair} & \textbf{0.72} & \textbf{0.79} & \textbf{0.75} & 0.91 & \textbf{0.39} & 1.07 & 0.84 & \textbf{0.14} & 1.18 \\ 
    \midrule
    \textit{short hair} & \textbf{0.32} & \textbf{0.46} & \textbf{0.28} & \textbf{0.66} & \textbf{0.46} & 1.30 & 0.94 & 1.07 & 1.73 \\ 
    \midrule
    \textit{straight hair} & \textbf{0.04} & \textbf{0.66} & \textbf{0.49} & \textbf{0.28} & 1.02 & \textbf{0.58} & \textbf{0.08} & 1.05 & \textbf{0.82} \\ 
    \midrule
    \textit{short} & \textbf{0.31} & 1.37 & 1.84 & 1.00 & \textbf{0.63} & 1.41 & 0.94 & 1.07 & 1.73 \\ 
    \midrule
    \textit{tall} & \textbf{0.17} & 1.01 & 1.18 & \textbf{0.00} & 1.01 & \textbf{0.72} & \textbf{0.79} & \textbf{0.49} & 1.18 \\ 
    \midrule
    \textit{dark skin} & \textbf{0.47} & \textbf{0.00} & 1.33 & 0.99 & \textbf{0.02} & 1.28 & \textbf{0.63} & \textbf{0.00} & 1.33 \\ 
    \midrule
    \textit{fair skin} & \textbf{0.15} & \textbf{0.16} & \textbf{0.21} & 0.98 & \textbf{0.13} & \textbf{0.87} & \textbf{0.36} & 1.41 & \textbf{0.81} \\ 
    \midrule
    \textit{fat} & \textbf{0.53} & \textbf{0.19} & \textbf{0.03} & 0.92 & \textbf{0.46} & \textbf{0.18} & \textbf{0.65} & \textbf{0.11} & \textbf{0.25} \\ 
    \midrule
    \textit{slim} & 0.88 & 1.48 & 1.55 & 0.98 & \textbf{0.63} & \textbf{0.64} & \textbf{0.24} & \textbf{0.78} & 1.02 \\ 
    \bottomrule
\end{longtable}
}
\twocolumn
\onecolumn

\bibliographystyle{splncs04}
\bibliography{main}

\end{document}